\pdfoutput=1
\documentclass[letterpaper, 10 pt, conference]{ieeeconf}  

\IEEEoverridecommandlockouts                              

\usepackage{times}
\usepackage{subcaption}
\usepackage{graphicx}

\usepackage{amsmath}
\usepackage{amssymb}
\usepackage{afterpage}
\usepackage{placeins}
\usepackage{booktabs}
\usepackage{siunitx}
\usepackage{float}

\usepackage[linesnumbered,ruled,vlined,commentsnumbered]{algorithm2e}
\usepackage{algorithmic}

\title{Autonomous Mobile Robot Navigation in Uneven and Unstructured  \\ Indoor Environments}

\author{Chaoqun Wang$^{*1,2}$, Lili Meng$^{*1}$, Sizhen She$^{1}$, Ian M.  Mitchell$^{1}$, Teng Li$^{1}$,  \\
Frederick Tung$^{1}$, Weiwei Wan$^{3}$, Max. Q. -H. Meng$^{2}$, and Clarence W. de Silva$^{1}$  
\thanks{$^{*}$ Indicates equal contribution.}
\thanks{$^{1}$Chaoqun Wang, Lili Meng, Sizhen She, Ian Mitchell, Teng Li, Frederick Tung and Clarence W. de Silva are with The University of British Columbia, Vancouver, BC, Canada.   {\tt\small \{lilimeng,desilva,tengli\}@mech.ubc.ca}, {\tt\small \{mitchell, ftung\}@cs.ubc.ca}}
\thanks{$^{2}$Chaoqun Wang, Max. Q. H. Meng are with The Chinese University of Hong Kong, China.
        {\tt\small \{cqwang, qhmeng\}@ee.cuhk.edu.hk} }
\thanks{ $^{3}$Weiwei Wan is from National Institute of Advanced Industrial Science and Technology, Japan.  {\tt\small wanweiwei07@gmail.com }  }
\thanks{}
} 

\begin{document}

\maketitle

\begin{abstract}
Robots are increasingly operating in indoor environments designed for and shared with people. However, robots working safely and autonomously in uneven and unstructured environments still face great challenges. Many modern indoor environments are designed with wheelchair accessibility in mind. This presents an opportunity for wheeled robots to navigate through sloped areas while avoiding staircases. In this paper, we present an integrated software and hardware system for autonomous mobile robot navigation in uneven and unstructured indoor environments. This modular and reusable software framework incorporates capabilities of perception and navigation. Our robot first builds a 3D OctoMap representation for the uneven environment with the 3D mapping using wheel odometry,  2D laser and RGB-D data. Then we project multi-layer 2D occupancy maps from OctoMap to generate the the traversable map based on layer differences. The safe traversable map serves as the input for efficient autonomous navigation. Furthermore, we employ a variable step size Rapidly Exploring Random Trees that could adjust the step size automatically, eliminating tuning step sizes according to environments. We conduct extensive experiments in simulation and real-world, demonstrating the efficacy and efficiency of our system. (Supplemented video link: https://youtu.be/6XJWcsH1fk0)

\end{abstract} 

\IEEEpeerreviewmaketitle 
\section{Introduction}

Autonomous mobile robot navigation plays a vital role in self-driving cars, warehouse robots, personal assistant robots and smart wheelchairs, especially with a shortage of workforce and an ever-increasing aging population. Significant progress has been achieved in recent decades  advancing the state-of-the-art of mobile robot technologies. These robots are operating more and more in unknown and unstructured environments, which requires a high degree of flexibility, perception, motion and control. Companies such as Google and Uber are developing advanced self-driving cars and expecting to present them into the market in the next few years. Various  mobile robots are roaming in factories and warehouses to automate the production lines and inventory, saving workers from walking daily marathons \cite{d2012guest}. Robot vacuums such as Roomba and Dyson360 eyes are moving around in the house to help clean the floors. Personal robots such as PR2 \cite{marder2010office,hornung2012navigation} and Care-O-bot \cite{reiser2013care} have demonstrated the ability to perform a variety of integrated tasks such as long-distance navigation and complex manipulation. 

\begin{figure}
	\begin{center}
\includegraphics[width =0.95\linewidth]{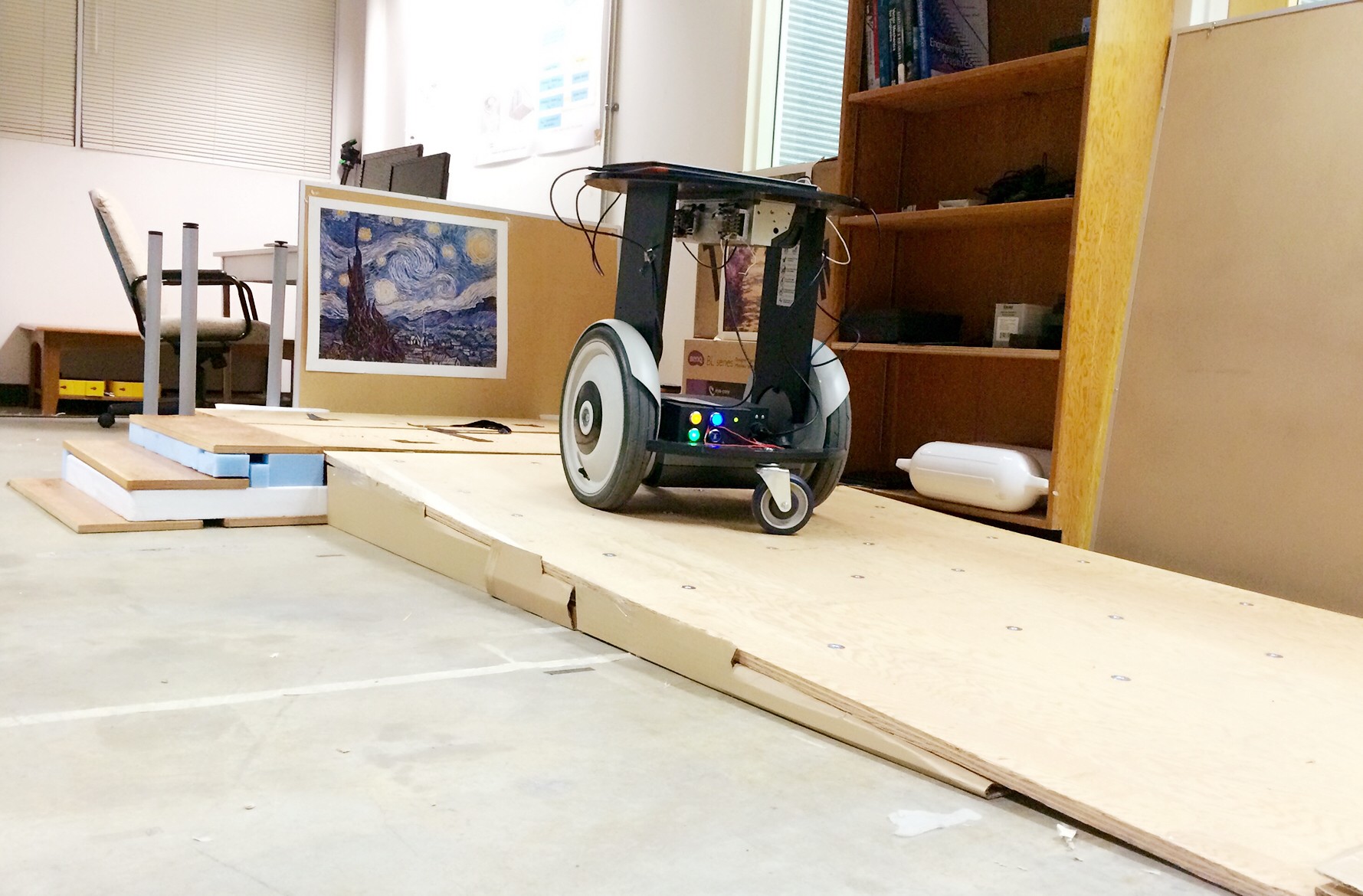}  
	\end{center}
	\vspace{-2mm}
	\caption{\textbf{The robot is navigating up the slope to the goal at the higher platform.} In the presence of staircases and slope, our robot first builds a 3D representation of the environment to generate the traversable map, and then the robot can navigate through the slope and avoid the staircases to reach the goal efficiently and safely.}
	\label{fig:real_robot}
    \vspace{0mm}
\end{figure}

\begin{figure*}[t]
	\begin{center}
		\includegraphics[width =0.9\linewidth]{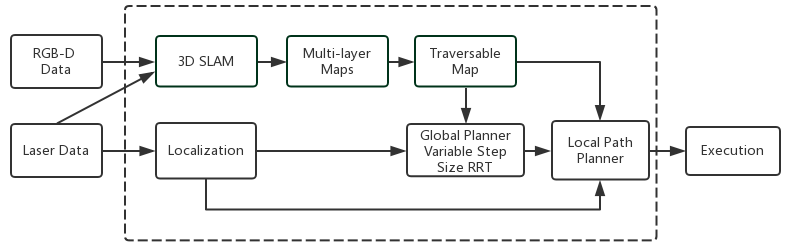}  
	\end{center}
	\vspace{-1mm}
	\caption{\textbf{High-level system architecture}. The robot first builds a 3D OctoMap representation for uneven environment with the present 3D SLAM using wheel odometry, a 2D laser and an RGB-D data. Multi-layer maps from OctoMap are used for generating the traversable map, which serves as the input for autonomous navigation. The robot employs a variable step size RRT approach for global planning,  adaptive Monte Carlo localization method to localize itself, and elastic bands method as the local planner to gap the global planning and real-time sensor-based robot control. }
	\label{fig:robot_software}
    \vspace{-2mm}
\end{figure*}

Mobile robots navigating autonomously and safely in uneven and unstructured environments still face great challenges.  Fortunately, more and more environments are designed and built for wheelchairs, providing sloped areas for wheeled robots to navigate through. However, little work focuses on an integrated system of autonomous navigation in sloped and unstructured indoor areas, especially narrow sloped areas and cluttered space in many modern buildings. The robots are required to safely navigate in narrow uneven areas such as those shown in Fig. \ref{fig:real_robot} while avoiding static and dynamic obstacles such as people and pets.

In this work, we present an integrated software and hardware framework for autonomous mobile robot navigation in uneven and unstructured indoor environments that are designed for and shared with people. Fig. \ref{fig:robot_software} shows a high-level system architecture of this work. Our robot first builds a 3D OctoMap representation for uneven environment with our 3D simultaneous localization and mapping (SLAM) using wheel odometry, a 2D laser and an RGB-D data. Then multi-layer maps are projected from OctoMap to generate the traversable map which serves as the input for our path planning and navigation. The robot employs a variable step size RRT approach for global planning, adaptive Monte Carlo localization method to localize itself, and elastic bands method as the local planner to close the gap between global path planning and real-time sensor-based robot control. Our focus is especially on efficient and robust environment representation and path planning. It is believed that reliable autonomous navigation in uneven and unstructured environments is not only useful for mobile robots but  also could provide helpful insight on smart wheelchair design in the future.

\section{Related work} 
Building autonomous robots that assist people in human environments has been a long-standing goal \cite{burgard1998interactive} and an appealing inspiration of research in robotics. Robots have demonstrated various levels of success but still face challenges. 

RHINO \cite{burgard1998interactive} integrates a distributed autonomous navigation architecture dedicated to human robot interaction with various sensors. Minerva \cite{thrun2000probabilistic} focuses on the probabilistic paradigm of environment, robot , sensors and models, setting the foundation for the probabilistic robotics epoch.  Besides autonomous navigation, Robox \cite{siegwart2003robox} integrated multi-modal interaction, speech recognition, and multi-robot coordination. Curious George \cite{meger2008curious} could use visual attention  locate the objects of interest in an environment. Jinny \cite{kim2004autonomous} can select the proper motion strategy according to different environments. Sonar or laser range finders are usually positioned on these robot to build 2D maps along a horizontal slice of the world \cite{thrun2005probabilistic}. However, just one slice of the space can not represent the environment, especially uneven environments with slopes and staircases. 

Three-dimensional (3D) environment sensing such as 3D Lidar which is on self-driving cars \cite{montemerlo2008junior}, 2D laser-range-finder with a pan-tilt unit \cite{zhang2014loam, zhang20133d}, or affordable RGB-D sensors is increasingly common. A 3D SLAM algorithm using 3D laser range finder is presented in \cite{cole2006using} which represents the world with point clouds, but neither free space nor unknown areas are modeled. Vision based 3D SLAM \cite{henry2014rgb, endres20143, Whelan14ijrr, mur2015orb} with affordable RGB-D sensors are increasingly popular since the introduction of KinectFusion \cite{newcombe2011kinectfusion}. These methods usually rely on iterative closest point registration (ICP) \cite{segal2009generalized} or local features matching based method or the integration of both. However, the ICP method is only confined to desk area \cite{newcombe2011kinectfusion}, while the local feature based methods are not robust in environments where white walls and dull floors dominate with few local features. Moreover, these methods only focuses on environment representation, without integrating autonomous navigation as an integrated system and demonstrating their overall performance. Autonomous navigation with a compact 3D representation of the environment is represented in \cite{marder2010office}, but only a layer of projected 2D environment representation is used for navigation. A combination of multi-layered 2D and 3D representations are used in \cite{hornung2012navigation} to improve planning speed, but only for an even environment. Robot autonomous navigation in full 3D environment is presented in \cite{maier2012real}, but it is designed for legged robots navigating through staircases. Moreover, the real-time interpreting and updating of 3D data still poses great challenges for low power laptops. These slow updates result in the robot either moving slowly or in a potentially unsafe manner. 

Some work focuses on uneven outdoor terrain. A rule-based fuzzy traversability index has been used \cite{howard2001rule} to quantify the ease-of-traversal of a terrain by a mobile robot based on measurements from image data. Online kernel-based learning is proposed to estimate a continuous surface over the area of interest while providing upper and lower bounds on that surface with 3D laser \cite{hadsell2009accurate}. However, these uneven outdoor terrains are very different from indoor uneven environments. Indoor environments tend to be mostly even, and often contain slopes for wheelchair accessibility. With robots or smart wheelchairs are operating in shared spaces indoors with people more and more, challenges operating in uneven and unstructured environments, especially in those designed for wheelchair mobility shall be addressed in the robotics community.

\section{The Robot Hardware and Software Platform}
\begin{figure}[t]
	\begin{center}
		\includegraphics[width =0.8\linewidth]{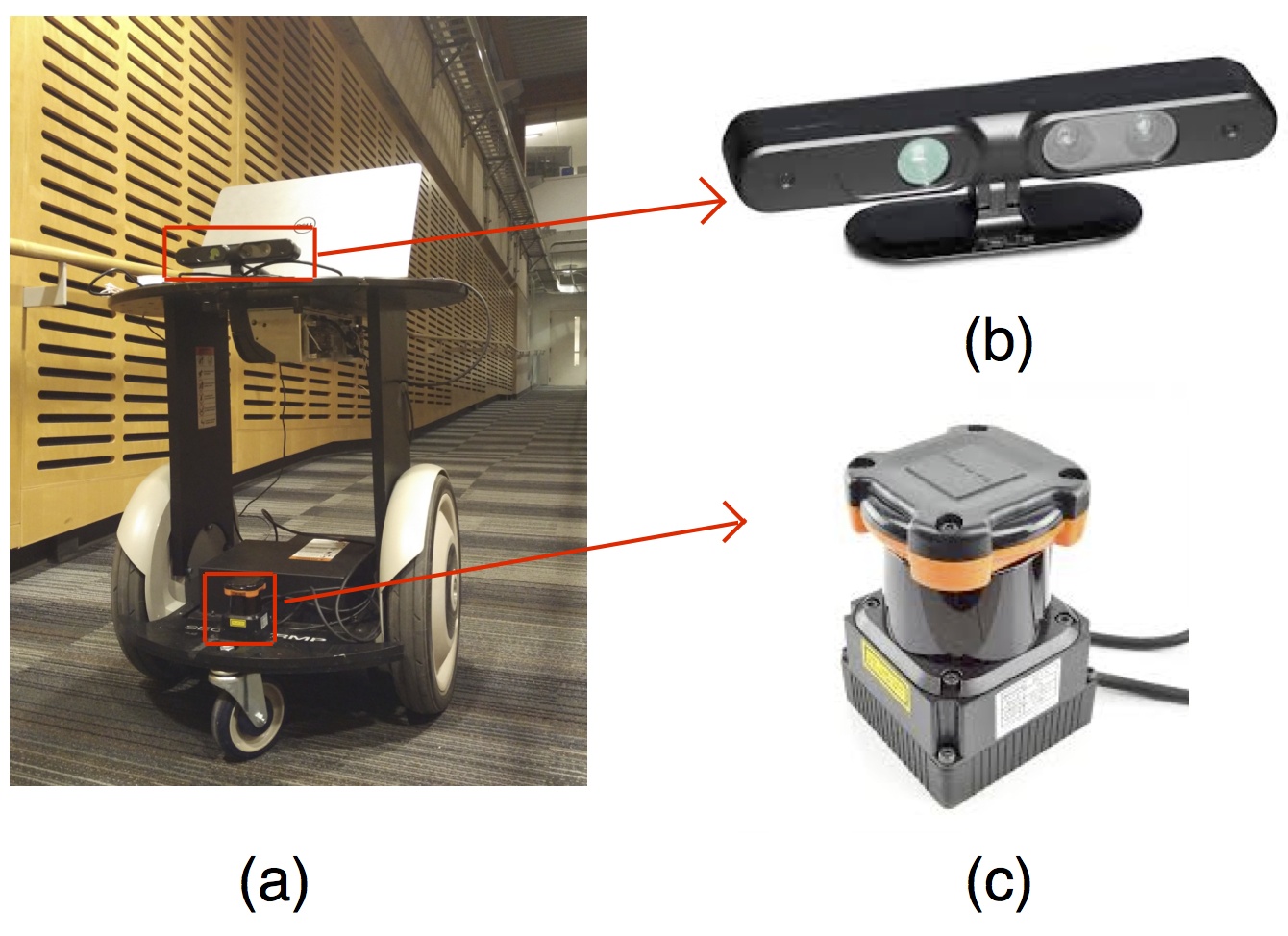}  
	\end{center}
	\vspace{-4mm}
	\caption{\textbf{Robot hardware platform.} (a) Segway robot in a sloped area. The robot base is segway RMP100 with custom installed casters for safety and onboard battery pack for providing power to sensors. (b) Xtion Pro Live
RGB-D camera is capable of providing 30Hz RGB and depth images, with 640x480 resolution and 58 HFV. (c) Hokuyo UTM-30LX laser scanner with  range 10m to 30m, and $270^{\circ}$ area scanning range for localization.}
	\label{fig:robot_hardware}
    \vspace{-2mm}
\end{figure}

The development of a mobile robot system to work around us and assist people is our long-term goal. The main design principle for the developed system is that each hardware and software component could work as both a single module and a part of an integrated system. To realize this principle, these hardware components are assembled using screws and adjustable frames, while the software platform uses the Robot Operating System (ROS) \cite{quigley2009ros}. 

Fig. \ref{fig:robot_hardware} shows the hardware platform for our robot. It includes a Segway RMP100 mobile base, an ASUS Xtion on the upper base, a Hokuyo UTM-30LX laser scanner mounted on the lower base, and a DELL laptop with Intel Core i7-4510U at 2GHz, 16GB memory (without GPU).

The software system is implemented in ROS Indigo release on top of an Ubuntu version 14.04LTS operating system. The 3D simulation experiments are performed on Gazebo \cite{koenig2004design}.  Fig. \ref{fig:robot_software} illustrates a high-level software architecture, and detailed key components of the software architecture will be described in Sec \ref{sec:environment_representation} and \ref{sec:planning}.

\section{Environment Representation}
\label{sec:environment_representation}
A consistent and accurate representation of the environment is a crucial component of autonomous systems as it serves as input for motion planning to generate collision-free and optimal motions for the robot. 

\subsection{3D mapping using wheel odometry, a 2D Laser and an RGB-D camera}
3D mapping pipelines commonly consist of localization and mapping components. Localization is the process to estimate robot pose, and mapping (or fusion) involves integrating new sensor observations into the 3D reconstruction of the environment. To overcome the challenge that vision based SLAM is not robust when the environment lacks local features, we employ wheel odometry, a 2D laser and an RGB-D camera concurrently to complement each other.

Our 3D Mapping framework builds on top of Karto SLAM \cite{kartoslam}, a 2D robust SLAM method containing scan matching, loop detection, Sparse Pose Adjustment \cite{konolige2010efficient} as the solver for pose optimization and 2D occupancy grid construction. Karto SLAM takes in data from the laser range finder and wheel odometry. It is the best performing ROS-enabled SLAM technique in the real world, being  less  affected by  noise \cite{santos2013evaluation} compared with other 2D SLAM methods, such as gmapping \cite{grisetti2007improved}, HectorSLAM \cite{kohlbrecher2013hector} and GraphSLAM \cite{thrun2005probabilistic}. 

\begin{figure}
 \centering
 \begin{minipage}{0.23\textwidth}
 \centering
 \includegraphics[width=\textwidth]{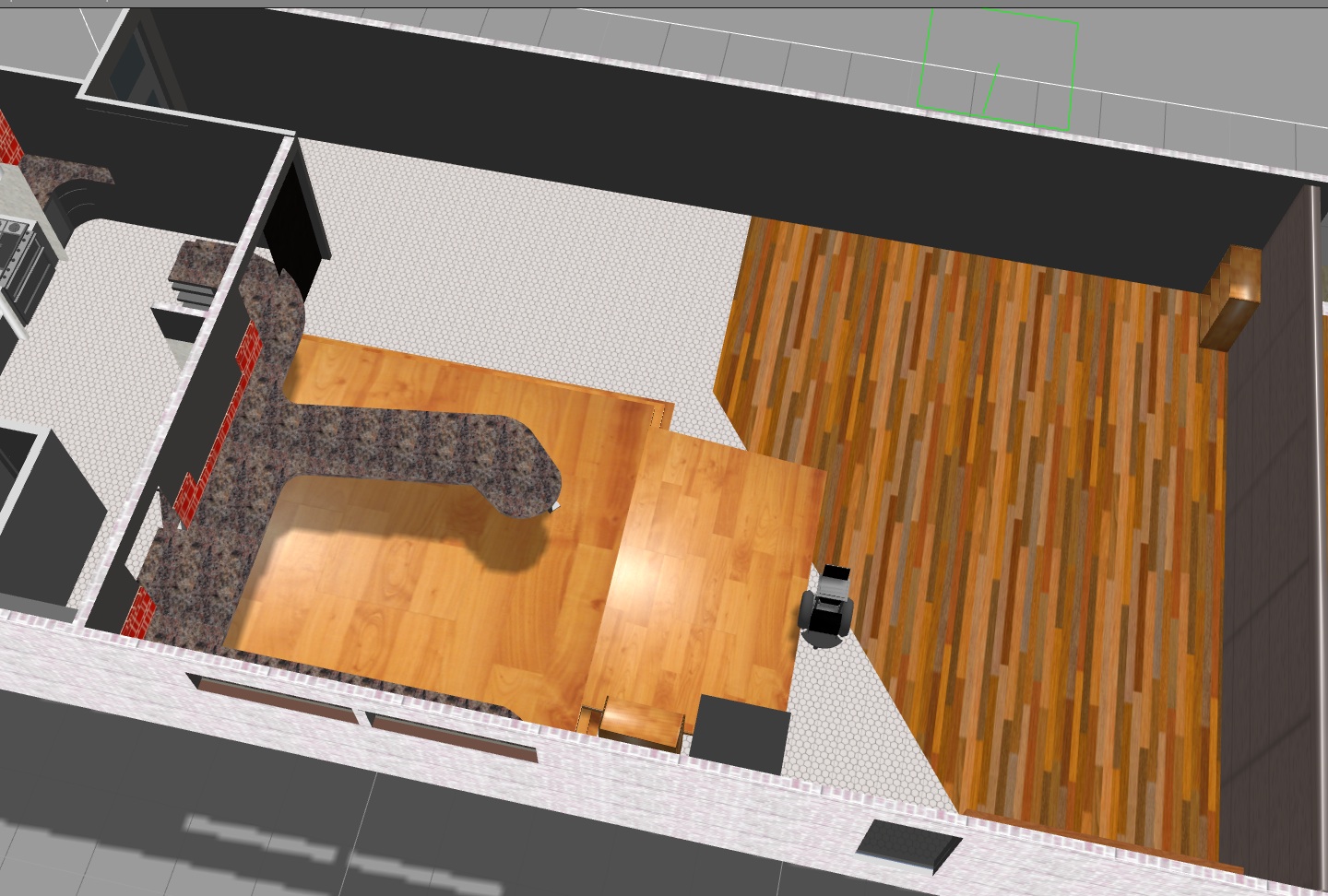}
 {(a) }
 \end{minipage} 
 \begin{minipage}{0.23\textwidth}
 \centering
\includegraphics[width=\textwidth]{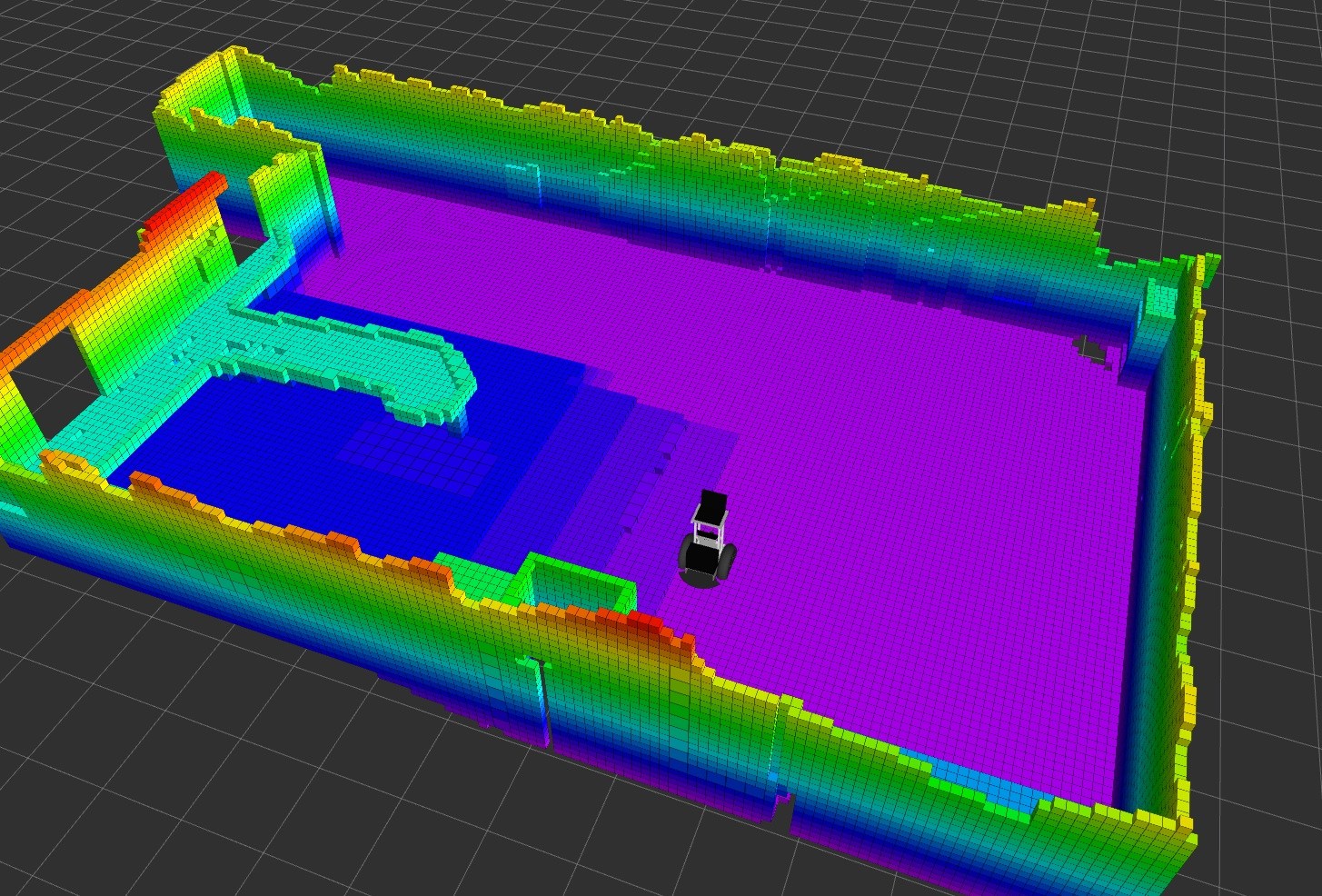}
 {(b) }
 \end{minipage} 
 \vspace{+0mm}
\caption{\textbf{3D environment representation of simulated world.} (a) Simulated environment model in gazebo  (b) 3D OctoMap environment built by our 3D SLAM using wheel odometry, a 2D laser scanner and an RGB-D sensor.}
\label{fig:environment}
\end{figure}

Instead of using Karto SLAM\rq s default 2D occupancy map,
which we found cannot represent uneven environment reliably, we build the environment based on OctoMap \cite{hornung2013octomap}. It is a probabilistic, flexible, and compact 3D mapping method which can represent the free, occupied and unknown environment. At each time step, the algorithm accepts point clouds of the environment using the RGB-D sensor and the localization information from the Karto SLAM using a 2D laser and wheel odometry. Fig. \ref{fig:environment}(b) shows an OctoMap representation of the simulated world generated by our 3D mapping. Note that free space is explicitly modeled in the experiment but is not shown in the figure for clarity.

\subsection{Multilayer maps and traversable map} 
\label{subsec:multilayer}


\begin{figure}
\centering
\begin{minipage}{0.5\textwidth}
\centering
\includegraphics[width=\textwidth]{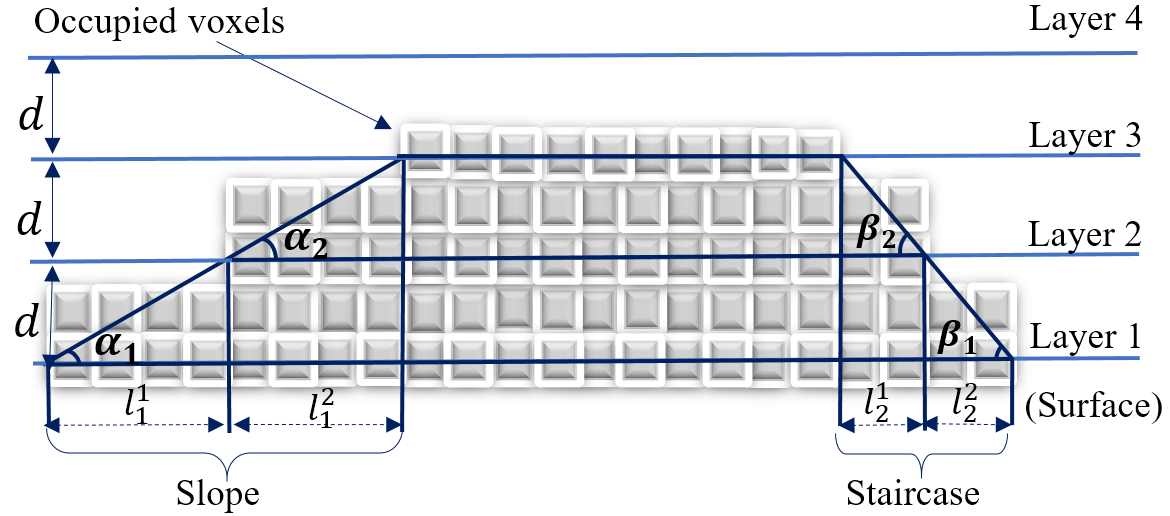}
{(a) }
\end{minipage} 
\begin{minipage}{0.5\textwidth}
\centering
\includegraphics[width=\textwidth]{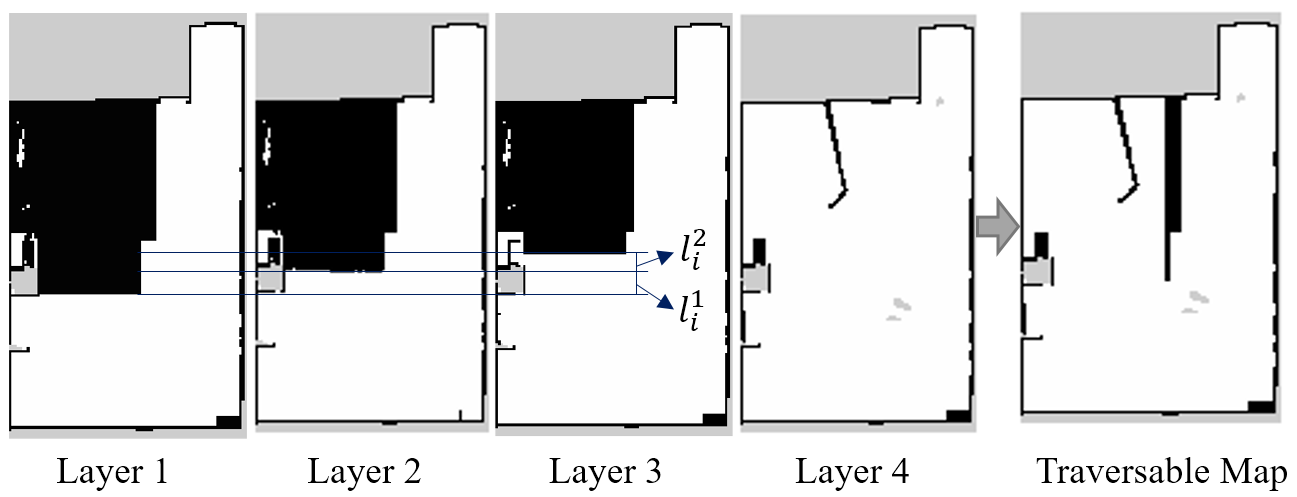}
{(b) }
\end{minipage} 
\caption{\textbf{Generation of traversable map from multilayer maps for the Gazebo Caffe environment.} (a) slope and staircase visualization with occupied voxels, (b) multi-layer maps and traversable map. In the traversable map, the staircases are occupied space, while the slope area except the edge is free space, which is safe for robot to navigate. For details please refer to Sec. \ref{subsec:multilayer}.}
\label{fig:multilayer maps}
\end{figure}
\vspace{0mm}
After the environment is represented by OctoMap as  mentioned above, the OctoMap is cut from bottom to top with several layers depending on the environment and the required accuracy. Then these multilayers are integrated into a traversable map, which is safe and efficient for the robot to navigate through. Both the multi-layer maps and the traversable map are represented as 2D grid occupancy maps, in which black represents occupied space, white as free space, and gray as unknown space.  A flag $F$ is used to mark the map traversablility and decide whether an area is traversable:
\begin{equation}
\label{eq:binary}
F=
\left\{
\begin{aligned}
0, & \quad \text{if} \ \alpha_i \geq  \theta, \quad \text{untraversable area}\\
1, & \quad \text{if} \ \alpha_i < \theta, \quad \text{slope, traversable area}
\end{aligned}
\right.
\end{equation}
where $\theta$ represents the angle threshold which can be set according to the robot climbing capability. $\alpha_i$ is the $i$th layer gradient:
\begin{equation}
 \alpha_i =arctan\frac{d}{l_i^j}=arctan\frac{d}{rv_i^j}
\label{equ:balance}
\end{equation}
where $d$ represents the distance between projected layers, as shown in Fig. \ref{fig:multilayer maps} (a). It can be set to be as small as the resolution of the OctoMap for accurate representation. $l_i^j$ represents the length of the edge difference between different layers, and it can be obtained through the number of voxels $v_i^j$ in that area and OctoMap resolution $r$. Take the environment in Fig. \ref{fig:environment} (a) as an example, first we project four layers of 2D map from the OctoMap as shown in Fig. \ref{fig:multilayer maps} (a) and the left four maps in Fig. \ref{fig:multilayer maps} (b). Then the gradient between layers are checked, for instance, in Fig. \ref{fig:multilayer maps} (a), both $\alpha_1$ and $\alpha_2$ are less than $\theta$, and that area is marked as a slope while $\beta_1$ and $\beta_2$ are larger than $\theta$, that area is marked as an untraversable area. At the same time, the left and right edges of the slope are marked as occupied for safety. Integrating all these multi-layer maps will generate the traversable map as shown in Fig. \ref{fig:multilayer maps} (b), which is very efficient for the mobile robot to use.   

\section{Planning and Navigation}
\label{sec:planning}
\begin{algorithm} [ht]
\KwIn{Start point $q_s$, Goal point $q_g$, $C_{free}$, $C_{obst}$ }
\KwOut{Path from $q_s$ to $q_g$  with several nodes}
\SetAlgoLined
 Initialization\;
 Establish a direct connection $l_{init}$ between $q_s$ and $q_g$ \; 
 \eIf{CollisionCheck($l_{init}$)}{
Return path $l_{init}$\;
 }{
 Get two points from $l_{init}$ as new start and end point \;  
 } 
 \While{iter $ \leq $ maxLoopNum $\&\&$ ConnectionFlag}{
 \For{$i={1,2}$}{
  $x_{rand}^{i} \leftarrow Sample(C_{free})$\;
  $x_{near}^{i} \leftarrow FindNearest(Tree^{i},x_{rand}^{i})$ \;
  $P_{temp} \leftarrow line(x_{near}^{i},x_{rand}^{i})$ \;
  \If{CollisionCheck($P_{temp}$)}{
   $x_{rand}^{i} $ $\leftarrow$ the point near obstacle \;
 } Add $x_{rand}^{i}$ to the current tree \;
 }
  \If{$x_{rand}^{1}=x_{rand}^{2}$}{
   Path $\leftarrow$ $ConnectTree(Tree^{1},Tree^{2})$ \;
   $ConnectionFlag=1;$
 }
 }
 \caption{Variable Step Size RRT }
\label{alg:variable_step_size_RRT}
\end{algorithm}
The traversable map serves as the input for the planning and navigation framework. We apply an efficient variable step size RRT planner as our global planner to construct a plan in a configuration space $\mathcal{C}$ to reach the goal.

\begin{figure}[b]
        \centering
        \begin{subfigure}[h]{0.23\textwidth}
            \centering
            \includegraphics[width=0.9\linewidth]{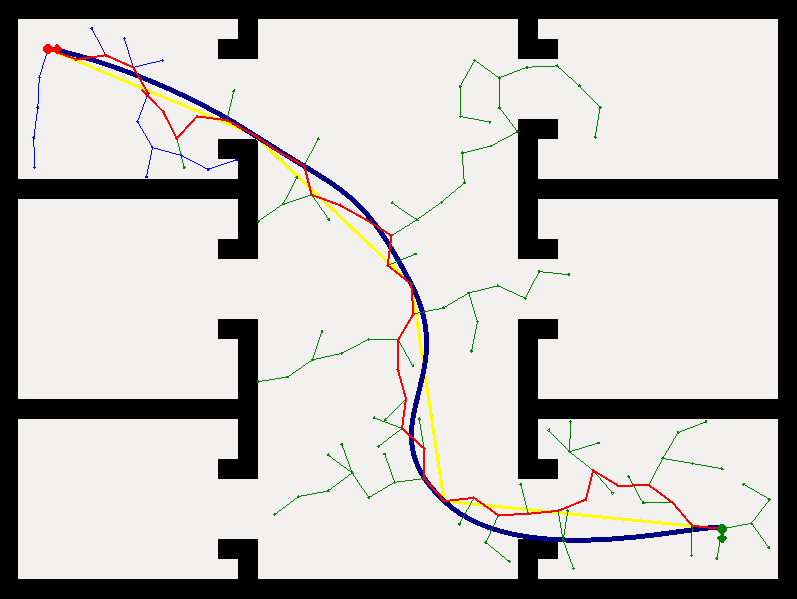}
       \caption[Step Size=10, Iter =768]%
            {{\small Step Size=30, Iter =171}}    
            \label{fig:mean and std of net14}
        \end{subfigure}
        \begin{subfigure}[h]{0.23\textwidth}  
            \centering 
            \includegraphics[width=0.9\linewidth]{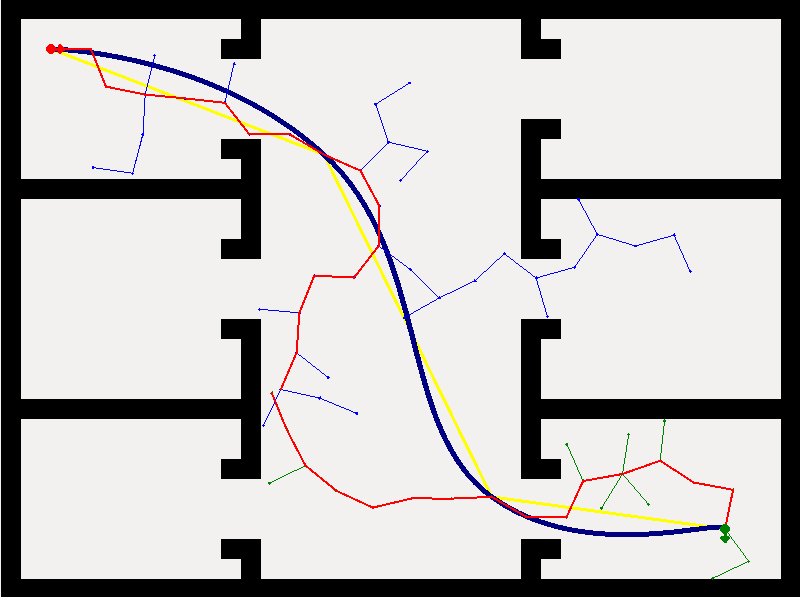}
            \caption[]%
            {{\small Step Size=40, Iter =57}}    
            \label{fig:mean and std of net24}
        \end{subfigure}
        \begin{subfigure}[h]{0.23\textwidth}   
            \centering 
            \includegraphics[width=0.9\linewidth]{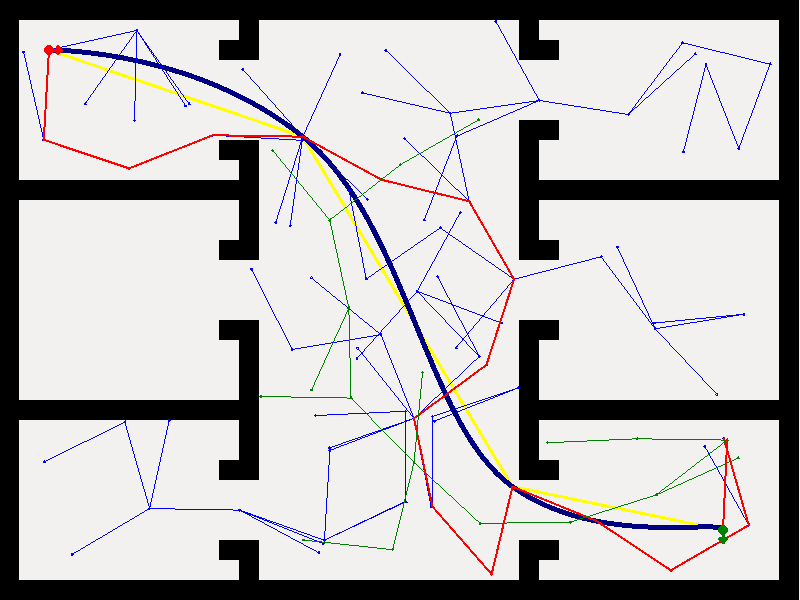}
            {{\small Step Size=90, Iter =166}}    
            \label{fig:mean and std of net34}
        \end{subfigure}   
        \begin{subfigure}[h]{0.23\textwidth}   
            \centering 
            \includegraphics[width=0.9\linewidth]{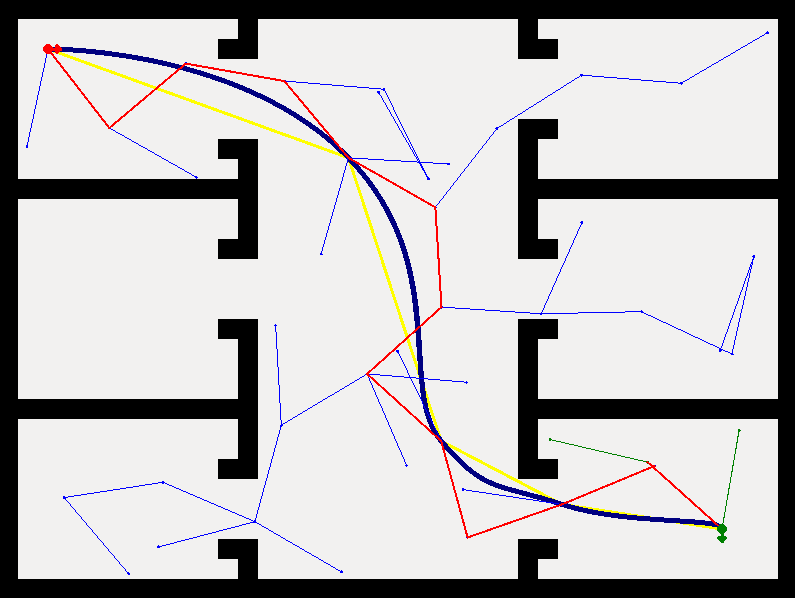}
            \caption[]%
            {{\small Step Size=Variable, Iter =73}}    
            \label{fig:mean and std of net44}
        \end{subfigure}
        \caption[ \textbf{Exploration with variable step size RRT} ]
        {\small \textbf{Comparison between RRT and our method.} (a)--(c) standard RRT. (d) our proposed variant step size RRT. The blue and green lines show the two trees rooted at start and goal point respectively. The red line is the sparse connection from start to goal. The dark blue line shows the optimal path following the red one.}
        \label{fig:exploration_variant_step_size}
\end{figure}

 The global planner generates a list of way-points for the overall optimal path from the starting position to the goal position through the map, while ignoring the kinematic and dynamic constraints of the robot. Then the local planner takes into account the robot kinematic and dynamic constraints, and generates a series of feasible local trajectories that can be executed from the current position, while avoiding obstacles and staying close to the global plan.

\subsection{Global planner: variable step size RRT}

\begin{figure*}
\centering
\begin{minipage}{0.19\textwidth}
\centering
\includegraphics[width=\textwidth]{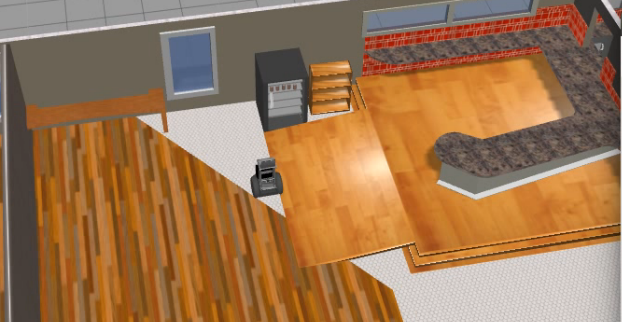}
\end{minipage} 
\vspace{+1.2mm}
\begin{minipage}{0.19\textwidth}
\centering
\includegraphics[width=\textwidth]{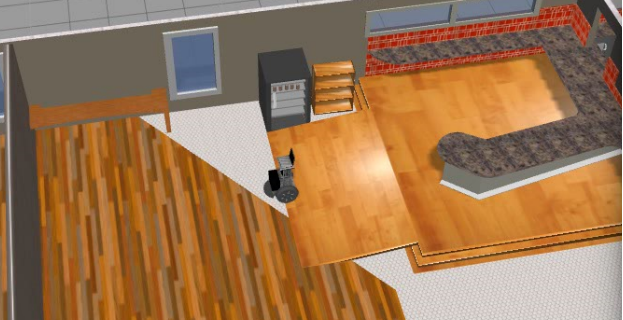}
\end{minipage} 
\vspace{+1.2mm}
\begin{minipage}{0.19\textwidth}
\centering
\includegraphics[width=\textwidth]{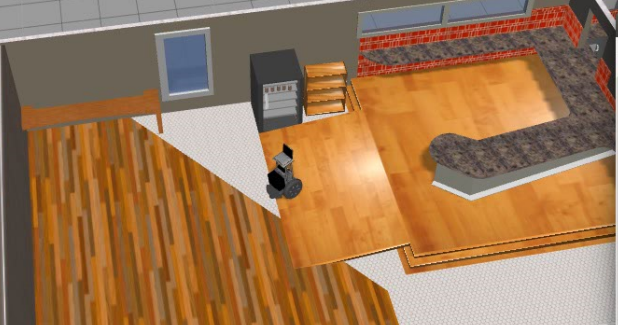}
\end{minipage}
\vspace{+1.2mm}
\begin{minipage}{0.19\textwidth}
\centering
\includegraphics[width=\textwidth]{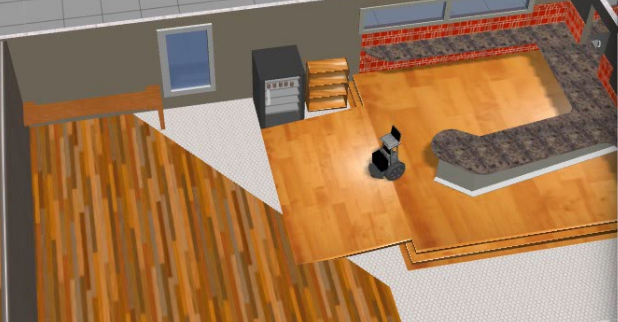}
\end{minipage} 
\begin{minipage}{0.19\textwidth}
\centering
\includegraphics[width=\textwidth]{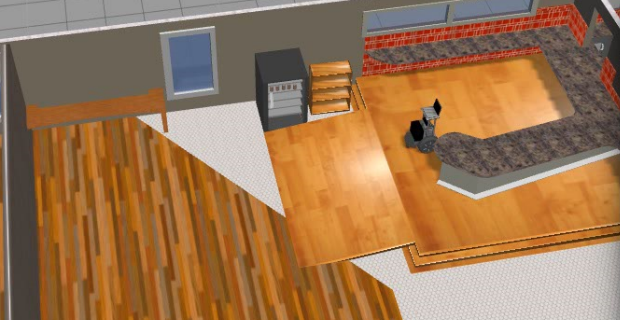}
\end{minipage} 
\begin{minipage}{0.19\textwidth}
\centering
\includegraphics[width=\textwidth]{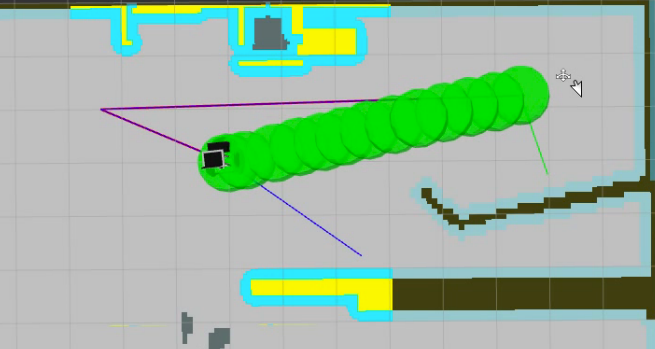}
{(a) }
\end{minipage} 
\begin{minipage}{0.19\textwidth}
\centering
\includegraphics[width=\textwidth]{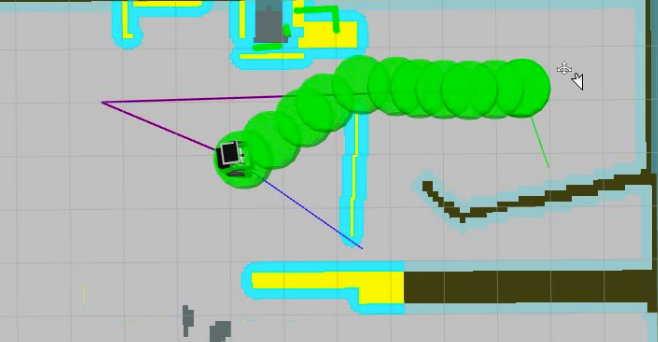}
{(b) }
\end{minipage} 
\begin{minipage}{0.19\textwidth}
\centering
\includegraphics[width=\textwidth]{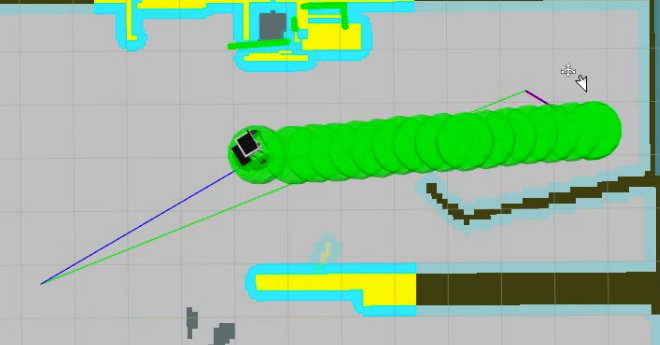}
{(c) }
\end{minipage}
\begin{minipage}{0.19\textwidth}
\centering
\includegraphics[width=\textwidth]{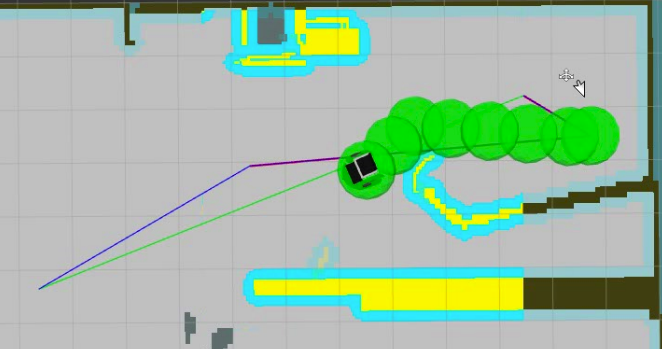}
{(d) }
\end{minipage} 
\begin{minipage}{0.19\textwidth}
\centering
\includegraphics[width=\textwidth]{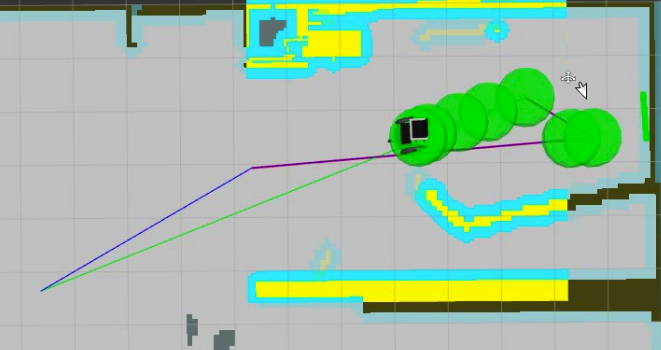}
{(e) }
\end{minipage}
\vspace{-1mm}
\caption{\textbf{Autonomous navigation in the simulated environment.} The first row shows images of autonomous navigation in Gazebo, and the second row shows the Rviz views of the process. For the whole process, please refer to our attached video. }
\label{fig:simulated_navigation}
\end{figure*}   
\vspace{0mm}

It is crucial that the global planning
process should be efficient so that the navigation system can run at a reasonable rate. Rapidly Exploring Random Trees (RRT) \cite{lavalle2001randomized} can handle problems with complex constraint and obstacles and perform well in high dimensional environments. It is good for our current setup and is also promising for integrating full 3D planning in the future. However, it needs to adjust the step size according to the environment, which is an inconvenient trial-and-error process. Therefore, we introduce a variable step size RRT for our global planning.  In the tree extending process of standard RRT, we get a sampling point $x_{rand}$ and then find the nearest point $x_{near}$ in the existing tree. If the segment that connects points $x_{rand}$ and $x_{near}$ is collision free, we move from point $x_{near}$ to $x_{rand}$ with limited step size $\sigma$, getting a new point $x_{new}$ that would be added to the existing tree.  In \cite{lavalle2000rapidly}, the extension process of RRT could be different for holonomic and nonholonomic systems. The latter can use the standard tree extending process while for the former, $x_{new}$ could be $x_{rand}$ in order to facilitate exploration. For a nonholonomic system, like the Segway robot we used in our experiment, it is important to choose the right $\sigma$. The RRT algorithm runs slowly when $\sigma$ is set small, i.e, with a small step size, it is slow to find the target. However, if $\sigma$ is too big, then the path generated by RRT will jitter, which is too long and inadequate for further trajectory optimization. Thus with standard extending process of RRT, it is vital to choose the proper $\sigma$ according to different environments. 
  
  According to our navigation framework that employs both global planner and local planner, the global path planner could be sub-optimal since the local planner would optimize the global path incrementally. Hence we propose to use a variable step size RRT that is originally used for holonomic systems. Our algorithm is illustrated in Algorithm \ref{alg:variable_step_size_RRT}. The main difference lies in the tree extending process. When a point $x_{nearest}$ is generated, instead of moving from the point to $x_{rand}$ with a limited step size, we connect the two points directly, and we get a line $l$. If $l$ passeses the collision checking process, then it is added to the tree directly. While if $l$ collides, the last point along the line would be $x_{new}$ and the point together with the line between $x_{new}$ and $x_{nearest}$ would be added to the tree. We recommend readers to read the our previous work \cite{wang2016variant} for more details. In addition, in order to deal with targets that lie in a narrow region, in which it will take longer before enough sampling points are chosen, we employ the idea of the RRT-connect algorithm \cite{kuffner2000rrt}, extending the tree from both the start and the goal position. The advantage of our method is that there is no need to adjust the parameter $\sigma$ in different environments. Whilst we try to extend the tree with the biggest step size, the planner can cover the configuration space with relatively small sampling points. Hence saving the time of collision checking of sampling points and finding the nearest point in the existing tree.
  
Fig. \ref{fig:exploration_variant_step_size} demonstrates the comparison between RRT of different step sizes and our proposed variable step size RRT. Fig. \ref{fig:exploration_variant_step_size} (a) shows that when the step size is small, it takes 171 iterations to reach the goal. When the step size is large such as 90 as shown in Fig. \ref{fig:exploration_variant_step_size} (c), it oscillates and takes 166 iterations to reach the goal. While our variable step size only takes 73 iterations to reach the goal. The standard RRT may outperform our method if the step size is adjusted properly, as shown in Fig.\ref{fig:exploration_variant_step_size}. However, in real implementation, our method does not need to adjust the step size in different environments and can achieve better comprehensive performance.

\subsection{Localization and local planner}
Mobile robot localization is the problem of determining the pose of a robot relative to a known environment \cite{thrun2005probabilistic}. The robot employs adaptive Monte Carlo localization (AMCL) \cite{fox1999monte}, a method that uses a particle filter to track the pose of the robot with a known map. It takes laser scans, the wheel odometry information and the traversable map, and then outputs pose estimation.

The local planner is seeded with the plan produced by the global planner, and attempts to follow it as closely as possible while taking into account the kinematics and dynamics of the robot as well as the obstacle information. We use Elastic Bands method \cite{quinlan1993elastic} as our local planner to close the gap between global path planning and real-time sensor-based robot control. 

During execution, the short and smooth path from the local planner is converted into motion commands for the robot mobile base. The local planner  computes  the  velocities  required  to
reach the next pose along the path and checks for collisions in the updated local costmap. 

\section{Experiments}
\label{sec:experiments}
Extensive experiments are conducted to evaluate the developed system both in simulation and real-world, demonstrating the efficacy for real-time mobile robot autonomous navigation in uneven and unstructured environments. It is notable that our focus is also on an integrated system besides the component methods.  

\subsection{Simulation experiments}
\begin{table}
\begin{center}
\begin{tabular}{lcccc}
\hline\noalign{\smallskip}
Example Tasks & Planning Time & Traveled Distance & Average Speed \\
\noalign{\smallskip}
\hline 
\noalign{\smallskip}
\multicolumn{4}{c}{\textbf{Simulated experiments}}\\
\noalign{\medskip}
Task-1 & 4.8$\pm$2ms & 8.6m & 0.67m/s \\
Task-2 & 5.9$\pm$2ms & 7.0m & 0.58m/s\\
\hline
\noalign{\medskip}
\multicolumn{4}{c}{\textbf{Real-world experiments}}\\
\noalign{\medskip}
Task-3 & 8.0$\pm$2ms & 8.3m & 0.66m/s   \\
Task-4 & 12.5$\pm$2ms & 9.0m & 0.38m/s  \\
Task-5 & 10.0$\pm$2ms & 7.0m & 0.35m/s \\
\noalign{\smallskip}
\hline
\end{tabular}
\caption{\textbf{Example tasks statistics of our robot operation.} Please refer to Sec \ref{sec:experiments} for more details of these representative tasks.}
\label{tab:statistics}
\end{center}
\vspace{-3mm}
\end{table}

The simulation is conducted in Caffe using an environment of size 13m x 10m x 3m, which we built in Gazebo. A visualization is shown in Fig. 4. The simulated robot model is equipped with wheel odometry, a 2D laser range finder and an RGB-D sensor. At the stage of 3D Mapping, the robot subscribes the point clouds from the RGB-D camera and the localization information from the 2D SLAM. Then the point clouds and localization information is fed to the Octomap server. The robot is tele-operated to move around the environment. Fig. \ref{fig:environment} (b) shows the OctoMap generated by our 3D Mapping. The resolution of the OctoMap is set as 0.05m to trade off between speed and accuracy, and the threshold on the occupancy probability is 0.7. Note that free space is explicitly modeled in the experiment but is not shown in the figure for clarity. 

The Octomap is then sliced and projected to four intersection planes as shown in the left part of Fig. \ref{fig:multilayer maps} (b). More planes could also be chosen depending on the accuracy necessity and the complexity of the environment. It can be seen that the staircase area shows steeper changes than the sloped area. Therefore, in the generation of the traversable map, we treat the steeper changes as untraversable area while the area with slower changes as the slope.  The right part of Fig. \ref{fig:multilayer maps} (b) shows the generated traversable map, which will be used for our robot autonomous navigation.

For autonomous navigation, we use the traversable map and sensors data as the input for path planning and localization. We conducted various autonomous navigation experiments in which the robot starts and reaches goals at different positions. Task-1 and Task-2 in Table \ref{tab:statistics} show some statistics of reaching the target on the ground and on higher platform respectively. Our variable step size path planner is very efficient, and only takes several milliseconds to reach the goal position.  Under the speed constraint of 1m/s, the robot can reach the target relatively fast. Fig. \ref{fig:simulated_navigation} illustrates an example of our robot autonomous navigation process in Task-2 in which the robot starts at lower ground and reaches the goal in the higher platform. The blue and green lines show the exploration process of our global planner. This process ended after a global path connecting the start point and the end point is found, shown in the red line. The local path planner can smooth the global path if it is not optimal, and the optimal path is marked as a concatenation of green bubbles. Fig.\ref{fig:simulated_navigation} (b) -- Fig.\ref{fig:simulated_navigation} (c) show the replanning process when an obstacle is found. If the height of a slope is higher than the laser on the robot, then the robot would consider there is an obstacle ahead. The robot can replan its path when this occurs.  Fig.\ref{fig:simulated_navigation} (d) -- Fig.\ref{fig:simulated_navigation} (e) show the replanning process of local planner when a real obstacle appears. The robot can avoid the obstacle and reach the target position.   

\begin{figure}
\centering 
\begin{minipage}{0.22\textwidth}
\centering
\includegraphics[width=\textwidth]{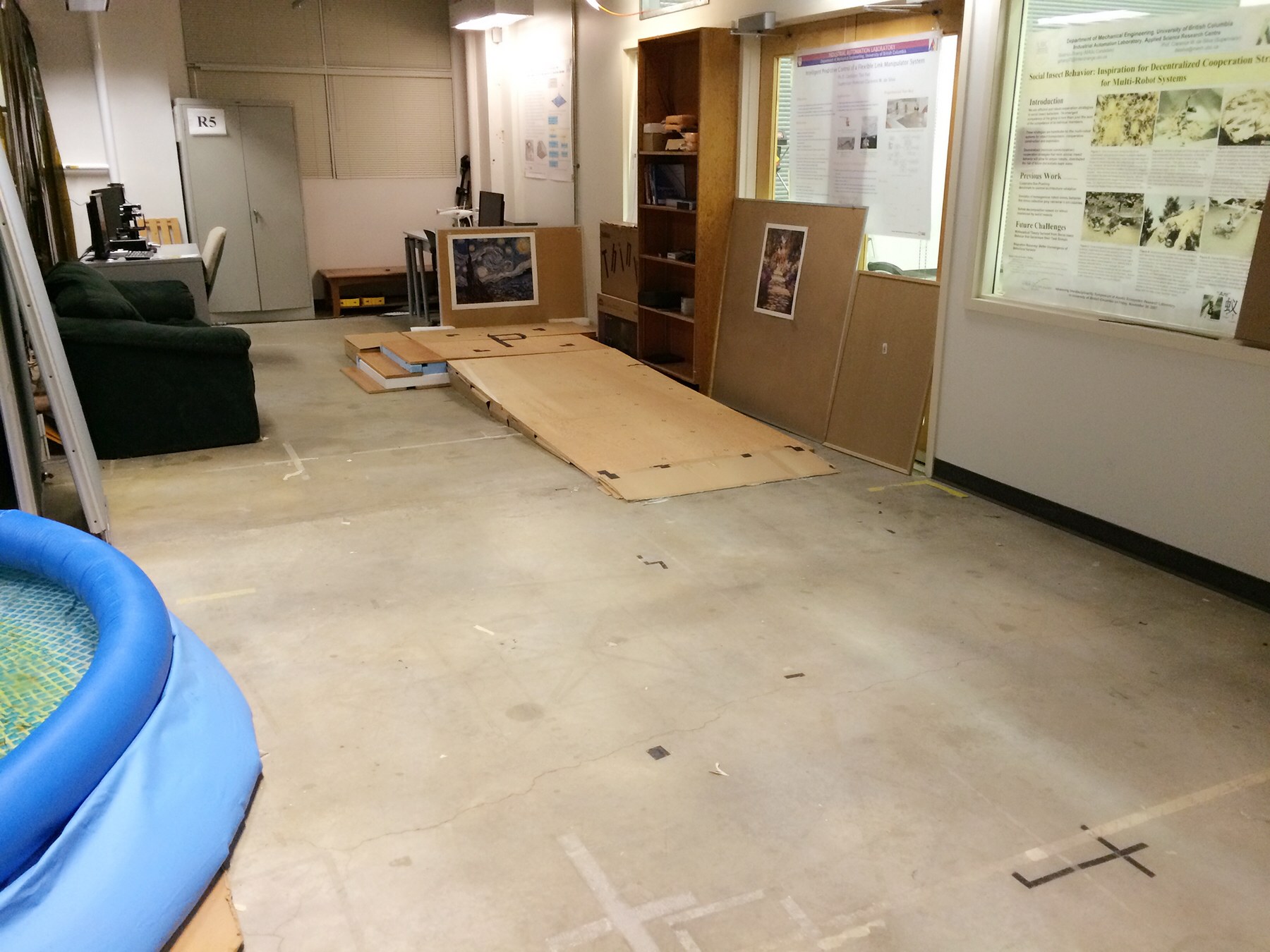}
{(a) }
\end{minipage} 
\begin{minipage}{0.25\textwidth}
\centering
\includegraphics[width=4.5cm,height=3cm]{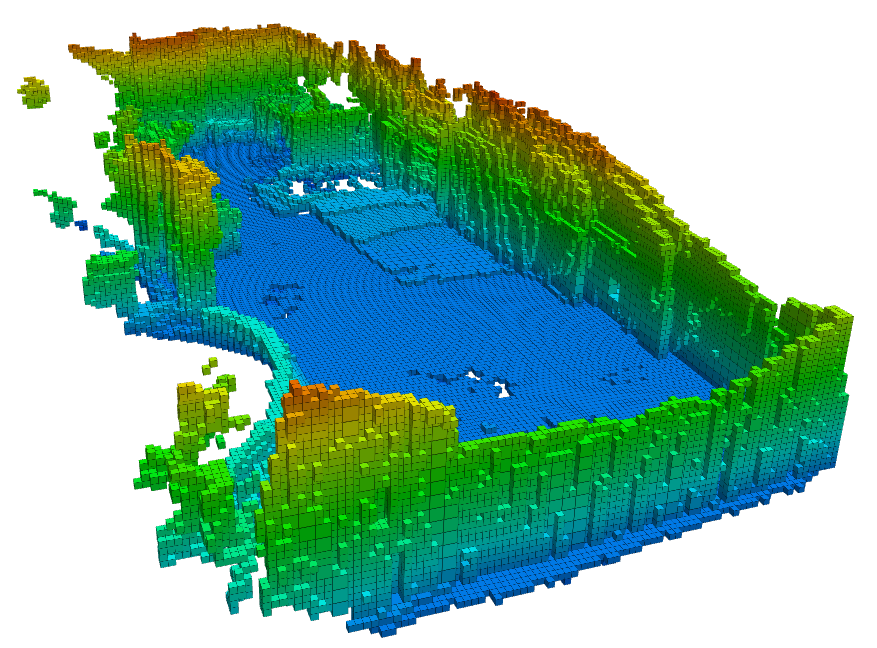}
{(b) }
\end{minipage}
\vspace{+1mm}
\caption{\textbf{3D mapping of a real environment.} (a) a photo of the real environment with spatial content of $11\times5\times3m$. (b) 3D representation of the environment with OctoMap. Only occupied voxels are shown for visualization.}
\label{fig:real_environment}
\vspace{-2mm}
\end{figure} 

\begin{figure*}
\centering
\begin{minipage}{0.19\textwidth}
\centering
\includegraphics[width=\textwidth]{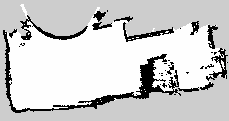}
{(a) }
\end{minipage} 
\begin{minipage}{0.19\textwidth}
\centering
\includegraphics[width=\textwidth]{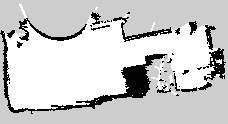}
{(b) }
\end{minipage} 
\begin{minipage}{0.19\textwidth}
\centering
\includegraphics[width=\textwidth]{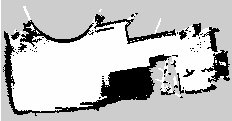}
{(c) }
\end{minipage}
\begin{minipage}{0.19\textwidth}
\centering
\includegraphics[width=\textwidth]{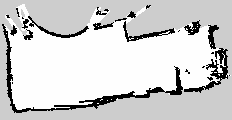}
{(d) }
\end{minipage} 
\begin{minipage}{0.19\textwidth}
\centering
\includegraphics[width=\textwidth]{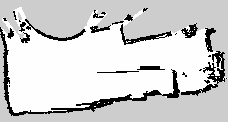}
{(e) }
\end{minipage} 
\vspace{-1mm}
\caption{\textbf{Multilayer maps and traversable map for real environment.} (a)-(d) multiple projected layers from OctoMap, (e) the traversable map. The staircases and slope edge are occupied while the slope is free space. }
\label{fig:multilayer_maps_real_environment}
\end{figure*}  
\vspace{0mm}

\begin{figure*}
\centering
\begin{minipage}{0.19\textwidth}
\centering
\includegraphics[width=\textwidth]{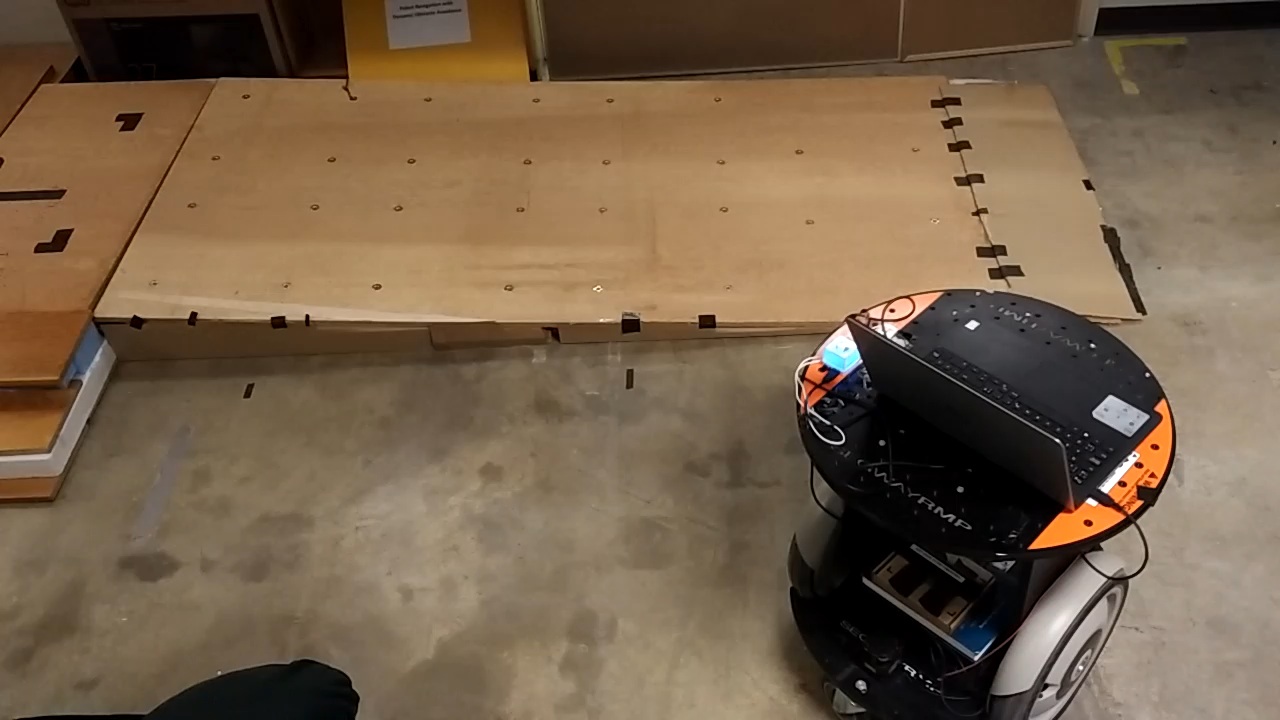}
\end{minipage} 
\vspace{+1.2mm}
\begin{minipage}{0.19\textwidth}
\centering
\includegraphics[width=\textwidth]{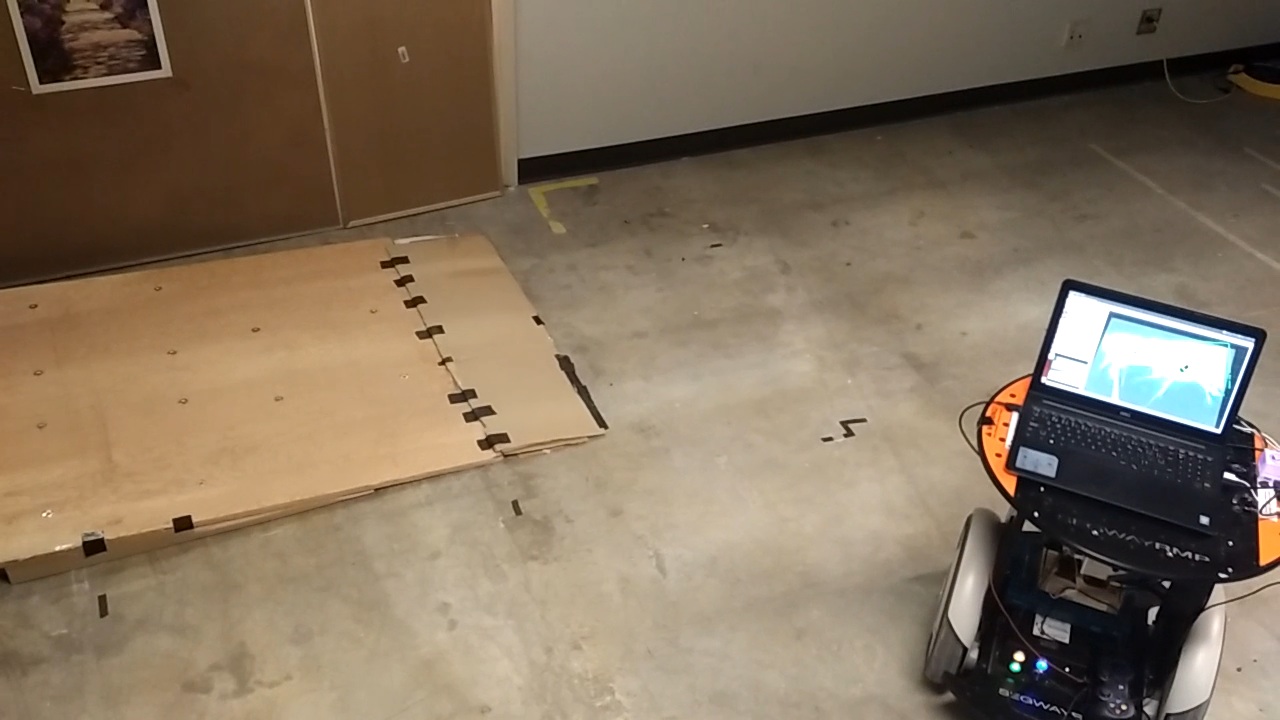}
\end{minipage} 
\vspace{+1.2mm}
\begin{minipage}{0.19\textwidth}
\centering
\includegraphics[width=\textwidth]{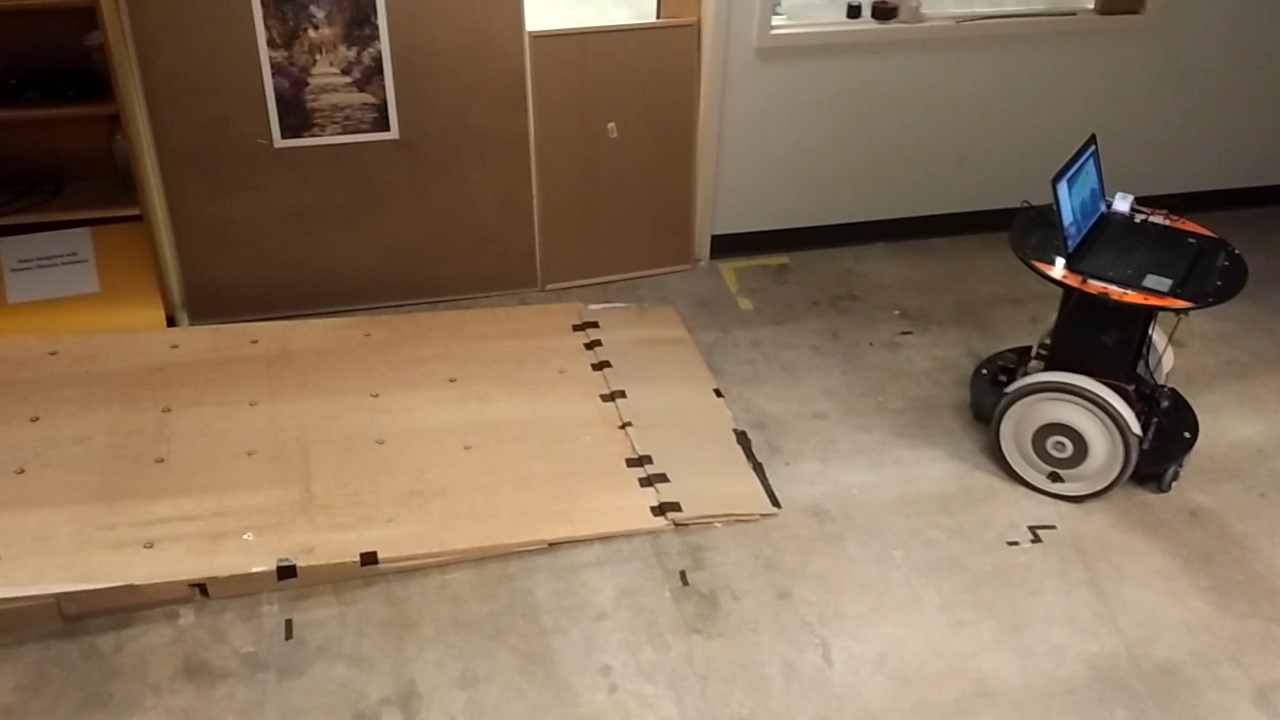}
\end{minipage}
\vspace{+1.2mm}
\begin{minipage}{0.19\textwidth}
\centering
\includegraphics[width=\textwidth]{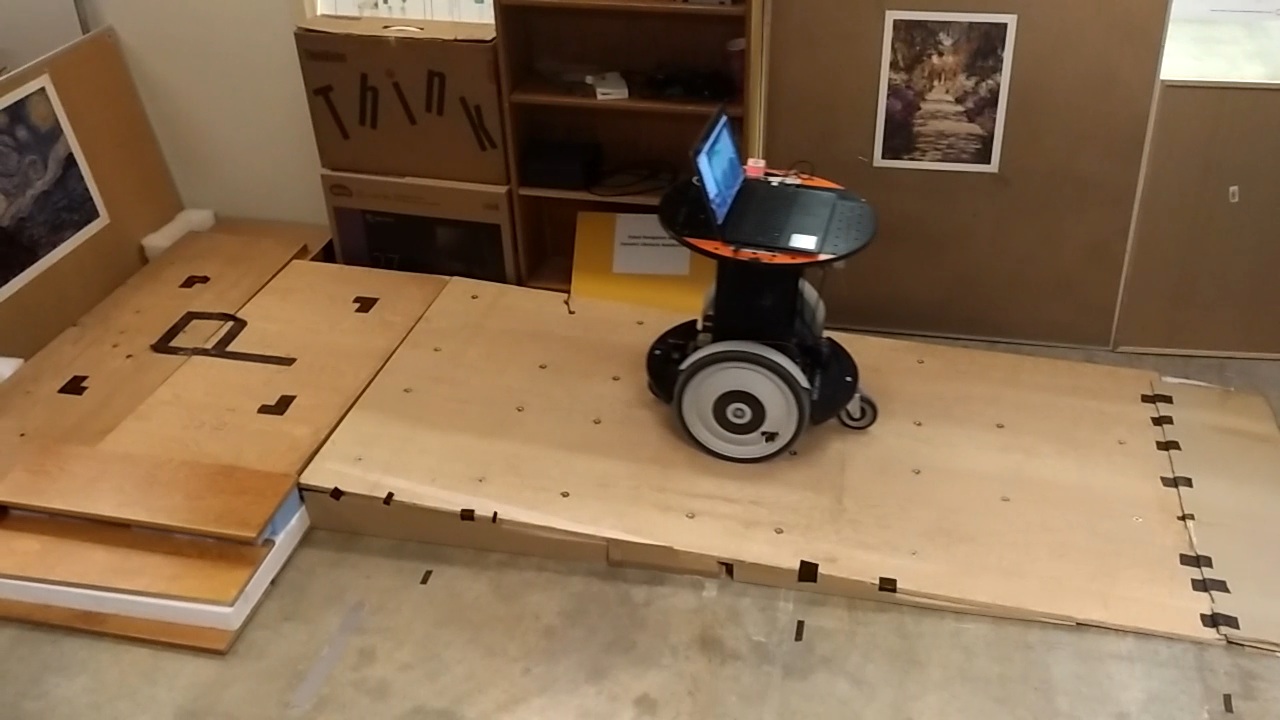}
\end{minipage} 
\begin{minipage}{0.19\textwidth}
\centering
\includegraphics[width=\textwidth]{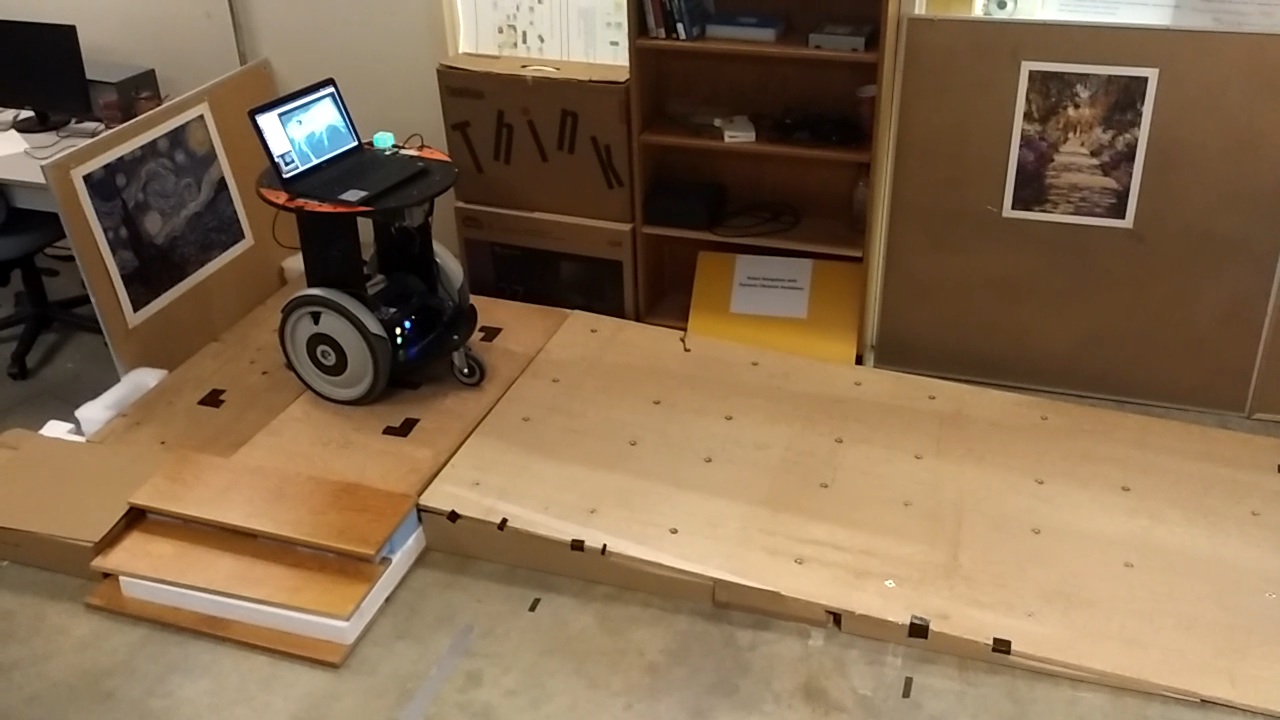}
\end{minipage} 
\begin{minipage}{0.19\textwidth}
\centering
\includegraphics[width=\textwidth]{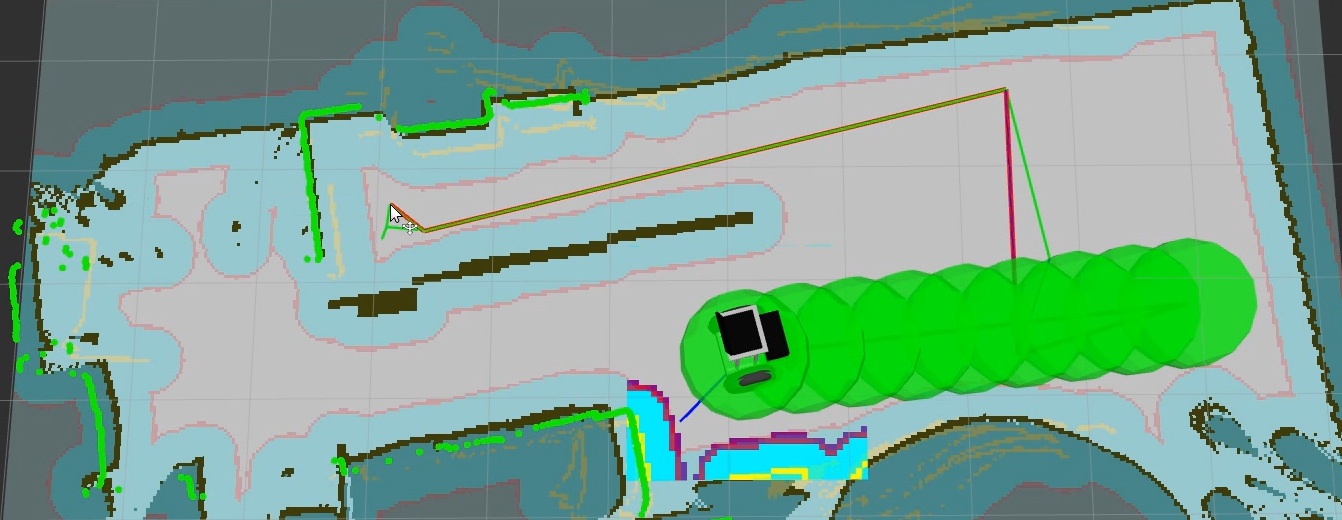}
{(a) }
\end{minipage} 
\begin{minipage}{0.19\textwidth}
\centering
\includegraphics[width=\textwidth]{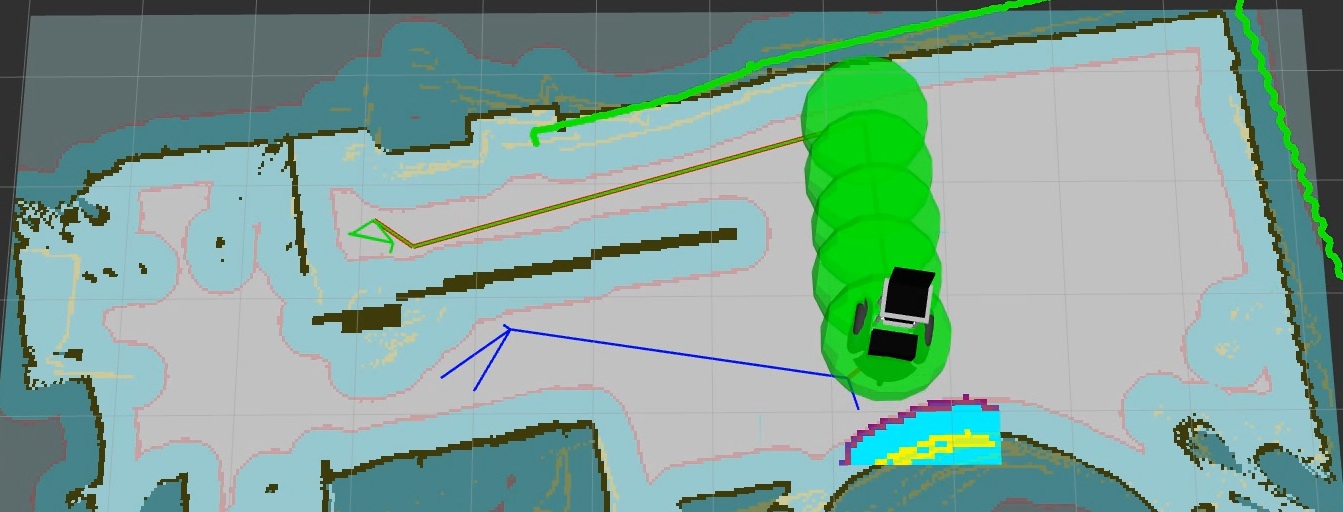}
{(b) }
\end{minipage} 
\begin{minipage}{0.19\textwidth}
\centering
\includegraphics[width=\textwidth]{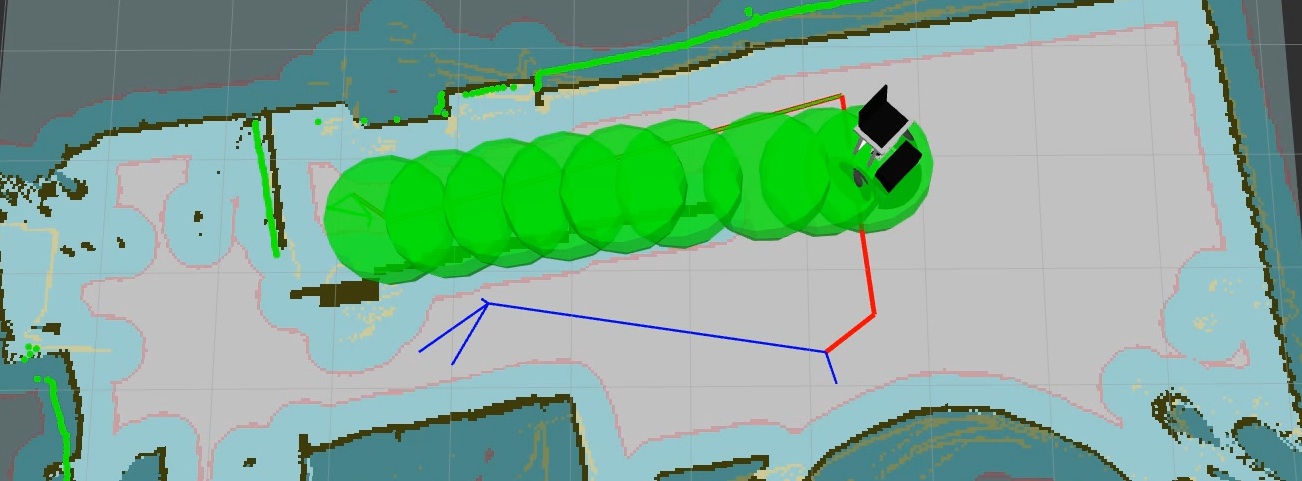}
{(c) }
\end{minipage}
\begin{minipage}{0.19\textwidth}
\centering
\includegraphics[width=\textwidth]{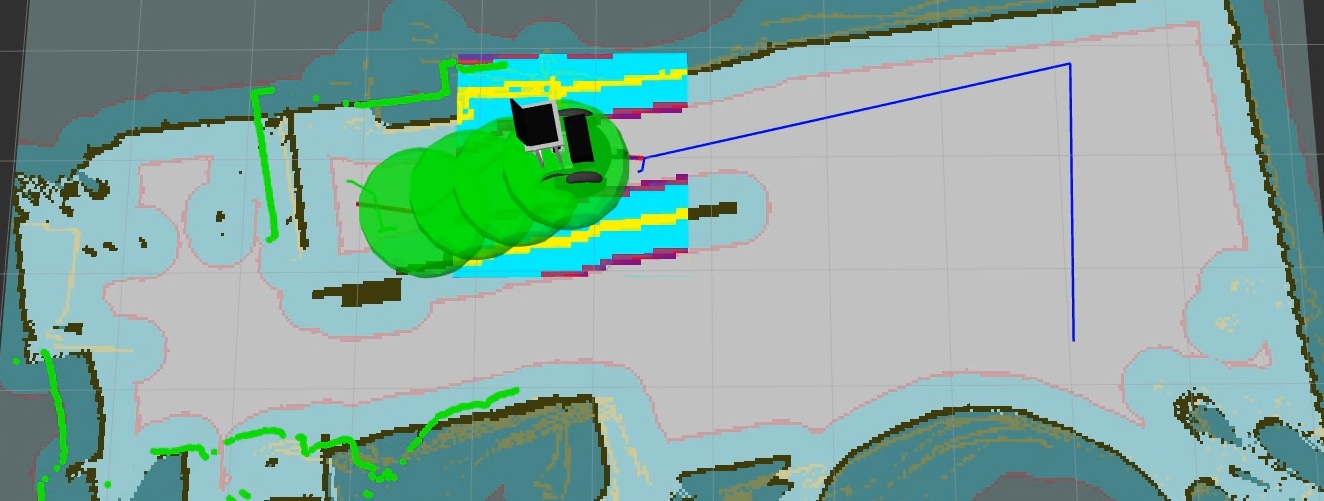}
{(d) }
\end{minipage} 
\begin{minipage}{0.19\textwidth}
\centering
\includegraphics[width=\textwidth]{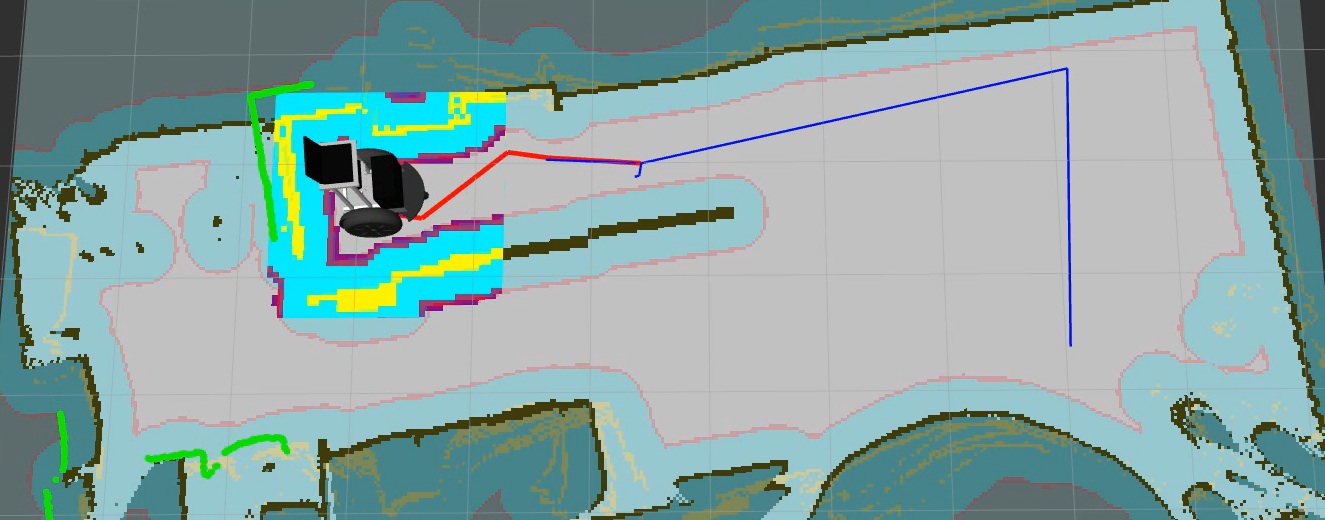}
{(e) }
\end{minipage}
\vspace{-1mm}
\caption{\textbf{ Robot autonomous navigation example in real environment.} The first row shows images of robot autonomous navigation, and the second row shows screenshots of Rviz views of the process. Our supplementary video shows more details.}
\label{fig:real_path_planning}
\end{figure*}    
\vspace{0mm}

\begin{figure}
\centering
\begin{minipage}{0.155\textwidth}
\centering
\includegraphics[width=\textwidth]{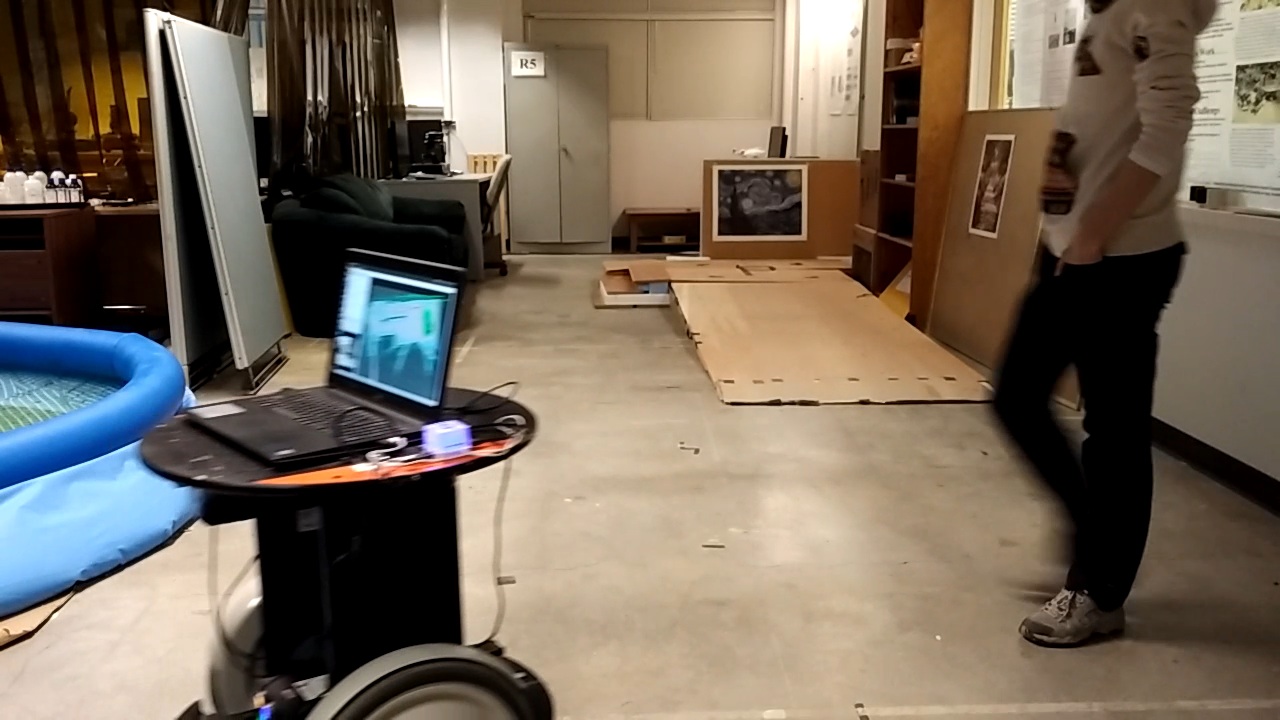}
\vspace{+1mm}
\end{minipage} 
\begin{minipage}{0.155\textwidth}
\centering
\includegraphics[width=\textwidth]{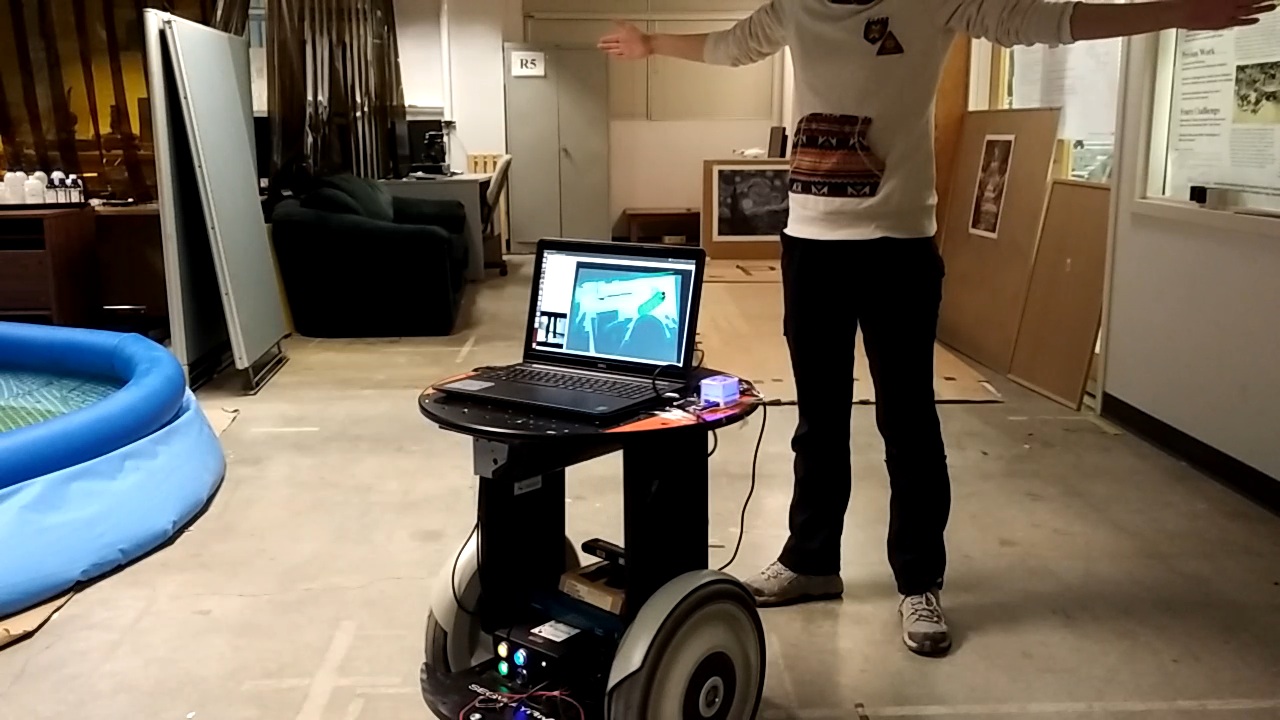}
\vspace{+1mm}
\end{minipage} 
\begin{minipage}{0.155\textwidth}
\centering
\includegraphics[width=\textwidth]{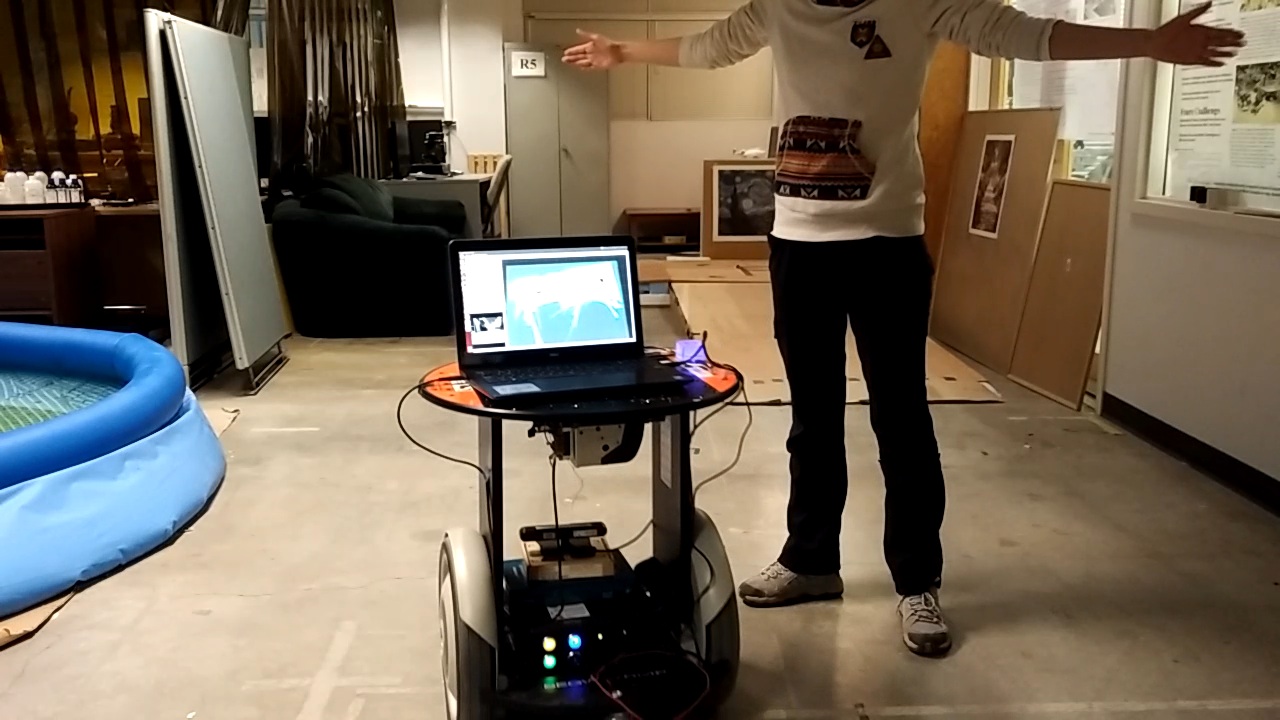}
\vspace{+1mm}
\end{minipage}
\begin{minipage}{0.155\textwidth}
\centering
\includegraphics[width=\textwidth]{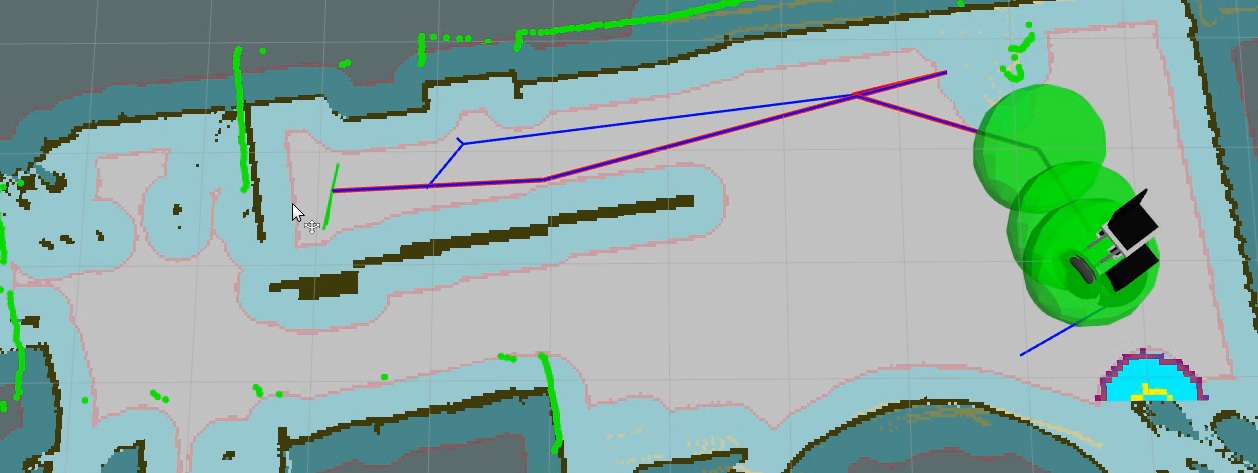}
{(a) }
\end{minipage} 
\begin{minipage}{0.155\textwidth}
\centering
\includegraphics[width=\textwidth]{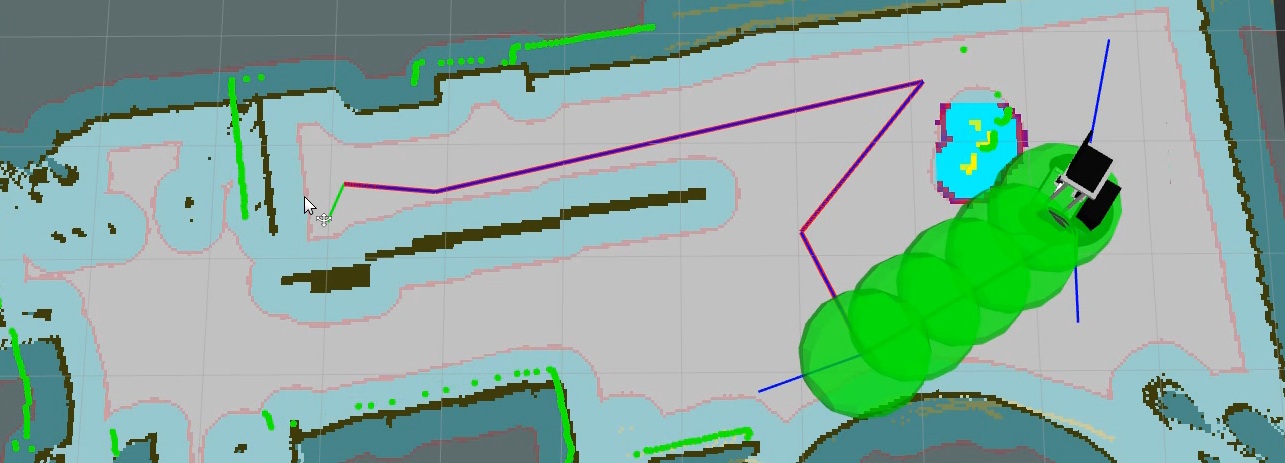}
{(b) }
\end{minipage} 
\begin{minipage}{0.155\textwidth}
\centering
\includegraphics[width=\textwidth]{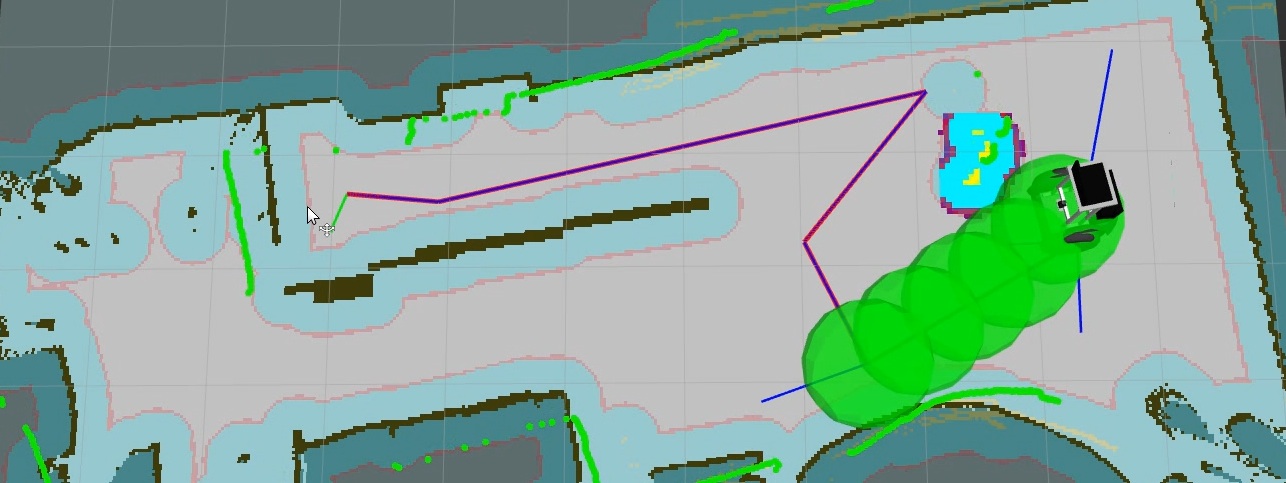}
{(c) }
\end{minipage}
\vspace{0mm}
\caption{\textbf{Dynamic obstacle avoidance.} (a) a dynamic obstacle is approaching the robot. (b) The human suddenly blocks the way in front of the robot. (c) The robot changes direction to avoid the human. Our supplementary video shows more details.}
\label{fig:dynamic_obstacle_avoidance}
\vspace{-4mm}
\end{figure}

\subsection{Real-world experiments}
We conduct extensive experiments with our Segway Robot in the real-world environment. Fig. \ref{fig:real_environment} shows our real environment with dimension of 11mx5mx3m, and the 3D OctoMap generated from our 3D mapping with wheel odometry, 2D laser and RGB-D data. At this 3D mapping stage, we teleoperate our Segway robot to move around the environment. Compared with the simulated environment, the real environment features challenges such as narrower slopes, different lighting conditions, and glass windows. The real robot is much more noisy due to long-term wear-and-tear, uncalibrated wheel odometry, and the disturbance of casters. The sensors are also noisier. Therefore, the OctoMap of the real environment and projected multi-layer maps in Fig. \ref{fig:multilayer_maps_real_environment} tend to have some distortions and inconsistent registrations compared with the simulated environment. All these noisy environment, robot and sensors demand robustness of our system.
The resolution of the OctoMap is experimentally set to 0.05m to trade off speed and accuracy. We also use four layers of projected maps to generate the traversable map in the real environment. The traversable map in Fig. \ref{fig:multilayer_maps_real_environment} (e) will serve as the input for path planning and localization in autonomous navigation. 

We present several representative tasks for the robot in our experiment in which all goal positions are located at higher platforms. In Task-3, the robot needs to navigate through the slope to reach the target from a far away starting position at ground level.  In Task-4, shown in Fig. \ref{fig:real_path_planning}, the robot starts near the slope but faces the staircase. With the traversable map, the robot could plan a path to avoid the staircase and navigate through the slope to reach the goal position.  Since the route contains many turns, the robot moves more slowly in this task. In Task-5, we test the dynamic obstacle avoidance ability of our system. A human suddenly appears and blocks the path of the robot, forcing the robot to replan and change directions to avoid the human. The average plan and replan time when encountering the obstacle is about 10ms. The robot can successfully avoid the obstacle and replan its path to the goal. In this dynamic obstacle case, the robot moves with a relatively low speed so as to safely avoid the obstacle. Fig. \ref{fig:dynamic_obstacle_avoidance} illustrates some of the procedures. 
Table \ref{tab:statistics} demonstrates the overall statistics of these representative tasks. 

One important parameter for robot navigating up the slope is the size of local costmap \cite{lu2014layered}. If the local costmap size is set too large, the far away ray sweep of the laser scan on the slope will intersect with the higher part of the slope, causing obstacle in the local costmap, which will block the robot's path. It is especially obvious in the more noisy real experiment in which the robot is prone to fail in localizing itself than in simulated environment. In case of localization failure, the robot will initiate the rotation recovery mode, in which the robot rotates and laser-sweeps in a circle.  If the local costmap size is too small, the obstacles may not be considered in time, which may cause unsafe collision. We experimentally determined the costmap of 3 meters width and 4 meters length to work best for the use in this case.

\section{Conclusion and Future Work} 
\label{sec:conclusion}

In this paper, we presented an integrated software and hardware architecture for autonomous mobile robot navigation in 3D uneven and unstructured indoor environments. This modular and reusable software framework incorporates capabilities of perception and navigation. We employ an efficient path planner which uses variable step size RRT in 3D environment with an octree-based representation. We generate a safe and feasible 2D map from multi-layer maps of 3D OctoMap for efficient planning and navigation. We demonstrate and evaluate our integrated system in both simulation and real-world experiments. Both simulation and real-robot experiments demonstrate the efficacy and efficiency of our methods, providing some insight for more autonomous mobile robots and wheelchairs working around us. Future work may integrate vision-based global localization \cite{backtrackingRegressionForest_IROS2017} \cite{LiliRandom}  and optimized path planning methods into our system. Moreover, it is promising to explore the possibility of using 3D semantic scene parsing \cite{tungmf3d} to understand the uneven areas and better differentiate between slopes and staircases. 

\section*{Acknowledgments}
The authors would like to thank Ji Zhang from Carnegie Mellon University for the helpful discussion and Wolfram Burgard from University of Freiburg for the suggestion on the supplementary video.

\bibliographystyle{unsrt}
\bibliography{references}
\end{document}